# The PBSAI Governance Ecosystem: A Multi-Agent AI Reference Architecture for Securing Enterprise AI Estates

John M. Willis
Quantum Powered Security Inc
jwillis@quantumpoweredsecurity.com
February 11, 2026

## Abstract

Enterprises are rapidly deploying large language models, retrieval-augmented generation pipelines, and tool-using agents into production, often on shared High-Performance Computing (HPC) clusters and cloud accelerator platforms that also support defensive analytics. These systems no longer operate as isolated models but as Artificial Intelligence (AI) estates: socio-technical systems spanning models, agents, data pipelines, security tooling, human workflows, and hyperscale infrastructure. Existing AI governance and security frameworks—such as the National Institute of Standards and Technology (NIST) AI Risk Management Framework (AI RMF), emerging AI regulations, and systems security engineering guidance (for example, NIST Special Publication (SP) 800-160 Volume 2)—articulate principles and functions but do not provide implementable architectures for multi-agent, AI-enabled cyber defense.

This paper introduces the Practitioner's Blueprint for Secure AI (PBSAI) Governance Ecosystem, a multi-agent reference architecture for securing AI estates at enterprise and hyperscale. PBSAI organizes responsibilities into a twelve-domain taxonomy and defines families of bounded agents that mediate between tools and policies, coordinated through a shared Model Context Protocol (MCP)-style context envelope and structured Output Contracts. The ecosystem assumes baseline capabilities for identity, telemetry, attestation, segmentation, resilience, policy enforcement, and evidence capture, and encodes key systems security techniques such as analytic monitoring, substantiated integrity, coordinated defense, and adaptive response. A lightweight formal model of agents and MCP envelopes, together with ecosystem-level invariants, clarifies the traceability, provenance, and human-in-the-loop guarantees the architecture enforces. We show how PBSAI aligns with NIST AI RMF functions (Govern, Map, Measure, Manage), supports regulatory objectives for high-risk AI systems, and applies in two illustrative scenarios: a medium-scale enterprise Security Operations Center (SOC) overlay and a hyperscale/HPC-backed defensive estate. PBSAI is offered as a structured, evidence-centric blueprint for an open ecosystem and as a foundation for reference implementations, empirical evaluation, and sector-specific overlays.

## Keywords

# 1 Introduction

Enterprises are rapidly deploying large language models (LLMs), retrieval-augmented generation (RAG) systems, tool-using agents, and traditional machine learning models into production. These systems no longer live as isolated artifacts. They are embedded in identity and access management, data pipelines, ticketing systems, security operations centers, and business workflows. In this paper, we refer to this socio-technical whole as an AI estate:

- models (LLMs, classical machine learning (ML), anomaly detectors);
- agents and orchestration layers;
- data pipelines and feature stores;
- security and monitoring tooling;
- human workflows and decision processes; and
- the underlying compute fabric, increasingly including High-Performance Computing (HPC) clusters, Graphics Processing Units (GPUs) and Tensor Processing Units (TPUs), and cloud-based accelerator infrastructures.

This same fabric underpins both production workloads and defensive capabilities. High-volume telemetry collection, real-time inference for detection, large-scale correlation, and replay-based hunting all depend on the same hyperscale infrastructure that serves customer-facing AI. As organizations increase their reliance on AI to detect, investigate, and respond to threats, the AI estate itself becomes a high-value target and a critical dependency for cyber defense (Sommer and Paxson, 2010; Tounsi and Rais, 2018; Weidinger et al., 2022).

## 1.1 Governance and security implementation gap

Over the last several years, governments, standards bodies, and industry consortia have produced a growing set of AI governance and security frameworks (OECD, 2019; Jobin et al., 2019; Floridi and Cowls, 2019; Batool et al., 2025; Papagiannidis et al., 2025). Examples include the NIST AI Risk Management Framework (NIST, 2023), national AI guidelines, emerging AI-specific regulations such as the EU Artificial Intelligence Act (European Union, 2024), and security-oriented standards such as NIST SP 800-160 Volume 2 and related control catalogs including NIST SP 800-53 (NIST, 2020, 2021; ISO/IEC, 2022, 2023). These documents articulate principles and functions such as:

- aligning AI with organizational values and risk appetite;
- ensuring transparency, robustness, and human oversight;
- managing data protection, supply-chain risk, and lifecycle security; and

- integrating AI into existing governance and risk management processes (NIST, 2021, 2023; OECD, 2019; Jobin et al., 2019).

However, these frameworks are largely agnostic about concrete system architectures. They say little about how to:

- structure an AI estate so that governance, security, and operations reinforce each other;
- coordinate multiple AI-enabled components (LLMs, agents, analytic models) across security domains;
- generate and preserve the evidence needed for auditability and accountability (Brundage et al., 2020); or
- make principled use of hyperscale (cloud-scale distributed) analytics and HPC resources for defense.

At the same time, security operations centers (SOCs) are built from a fragmented assortment of commercial tools: Security Information and Event Management (SIEM), Endpoint Detection and Response (EDR), User and Entity Behavior Analytics (UEBA), Security Orchestration, Automation, and Response (SOAR), cloud security, identity and access management, data protection, and more. Each vendor is now adding AI features in its own way—LLM "assistants," anomaly detectors, automated playbooks—despite existing standards for structured threat intelligence exchange (e.g., STIX/TAXII), there is no shared architectural pattern or interoperable evidence model for AI-generated analytic outputs (Sommer and Paxson, 2010; Tounsi and Rais, 2018). This creates a risk that AI for cyber defense will emerge as a patchwork of opaque, vendor-specific capabilities that are difficult to govern, validate, or reason about at the estate level (Batool et al., 2025; Papagiannidis et al., 2025).

**1.2 From single models to multi-agent defensive ecosystems**

Most existing discussions of "AI security" and "AI for security" are still framed at the model or tool level:

- how to secure and test an individual model;
- how to use an LLM to summarize alerts;
- how to apply anomaly detection to specific telemetry streams (Sommer and Paxson, 2010; Weidinger et al., 2022).

These are important questions, but they do not address how to secure the AI estate as a whole, especially when:

- multiple models and agents are interacting over shared data and infrastructure;
- security controls and detections are distributed across dozens of tools;

- governance and compliance teams need a coherent view of risk and evidence (Jobin et al., 2019; Floridi and Cowls, 2019); and
- the estate is partially or wholly dependent on hyperscale compute platforms.

In practice, what SOCs and governance bodies need is not another point solution, but a multi-agent defensive ecosystem that:

1. sits as a tool-agnostic management overlay above existing SIEM, EDR, Identity and Access Management (IAM), SOAR, cloud, and data platforms;
2. coordinates families of agents across governance, monitoring, incident response, resilience, supply-chain, and architecture domains;
3. uses a consistent agent abstraction and context envelope so that decisions and actions are traceable and subject to policy;
4. produces a structured evidence graph that links controls, events, models, agents, and governance decisions (Brundage et al., 2020); and
5. can exploit HPC and cloud accelerators for large-scale analytics, hunting, and model validation, without sacrificing oversight.

This paper proposes such an ecosystem in the form of the PBSAI Governance Ecosystem, a reference architecture for securing AI estates using coordinated multi-agent patterns.

**1.3 Core idea: evidence-centric, governance-aligned multi-agent architecture**

The central design choice in PBSAI is to treat governance, security, and AI enablement as a single systems engineering problem, consistent with cyber-resilient systems guidance (NIST, 2021) and emerging AI governance literature (Batool et al., 2025; Papagiannidis et al., 2025). Rather than bolting AI onto existing tools or treating AI governance as a purely policy-level exercise, PBSAI:

- organizes the security and governance landscape into a twelve-domain taxonomy (Governance, Risk, and Compliance (GRC), asset and configuration, identity, monitoring, protection, data security, incident response, resilience, architecture, physical security, supply chain, and program enablement);
- introduces agent families in each domain—small software workers between tools and policies—that ingest structured events, apply deterministic logic plus bounded LLM assistance, and emit signed, schema-constrained outputs;
- standardizes an MCP-style context envelope that carries mission, thread, task, policy references, constraints, decision basis, provenance, and classification/legal fields with every agent invocation; and
- defines a minimal secure AI stack and baseline (identity and access, logging and telemetry, attestation, segmentation, resilience, governance policies, and evidence/schema registries) as prerequisites for safe deployment (NIST, 2020, 2021, 2023; CISA, 2020; OMB, 2018).

By combining these elements, PBSAI aims to provide a concrete reference architecture that organizations can adapt, implement incrementally, and map directly onto existing AI governance and systems security frameworks (NIST, 2021, 2023; OECD, 2019; European Union, 2024; ISO/IEC, 2023).

### 1.4 Contributions

This paper makes four main contributions.

1. **PBSAI Governance Ecosystem.**
   We introduce a twelve-domain, multi-agent defensive architecture for AI estates. Each domain hosts agent families that implement specific security and governance responsibilities, coordinated via shared schemas and MCP-style context envelopes. The ecosystem is rooted in systems security engineering concepts—for example, NIST SP 800-160 Volume 2 techniques for cyber-resilient systems—and is informed by high-value asset assessment practices in government and critical-infrastructure contexts (NIST, 2021; CISA, 2020; OMB, 2018).

2. **Minimal Secure AI Stack and Baseline.**
   We define a minimal set of technical and governance assumptions required before deploying AI-enabled agents into critical cyber workflows. These include identity and access foundations, centralized telemetry, platform and agent attestation, segmentation and zero trust posture, backup and continuity, policy and control libraries, human-in-the-loop thresholds, Crown Jewel inventories, and evidence and schema registries (NIST, 2020, 2021; ISO/IEC, 2022). For hyperscale environments, we extend the baseline to include consistent attestation and integrated telemetry across multi-cluster HPC and cloud estates, and capacity planning for AI-assisted defense workloads.

3. **Agent Design Pattern and MCP-Style Envelope.**
   We specify a reusable agent pattern for AI-for-cyber: event-driven workers that apply deterministic logic first, invoke LLMs only where bounded judgment or language understanding is needed, and emit signed Output Contracts under a common MCP-style context wrapper. The envelope explicitly encodes mission, intent, policy references, constraints, decision basis, and provenance, enabling traceability, auditability, and policy routing across the ecosystem (Brundage et al., 2020).

4. **Alignment with NIST AI RMF, NIST SP 800-160v2, and Regulatory Objectives.**
   We show how PBSAI maps onto NIST SP 800-160 Volume 2 techniques and approaches—for example, Analytic Monitoring, Substantiated Integrity,

Coordinated Defense, and Adaptive Response—and the NIST AI Risk Management Framework functions (Govern, Map, Measure, Manage) (NIST, 2021, 2023). We also show how the ecosystem can support regulatory obligations for high-risk AI systems and sectoral regimes, including data governance, human oversight, robustness, incident reporting, continuity, and supply-chain assurance (OECD, 2019; European Union, 2024; Papagiannidis et al., 2025). We illustrate this using two application scenarios: a medium-scale enterprise SOC using PBSAI as a multi-agent overlay on its existing stack, and a hyperscale/HPC-backed organization using PBSAI to coordinate streaming analytics, large-scale replay and hunting, AI validation, and constrained automated containment.

Taken together, these contributions aim to move the discussion of "secure and trustworthy AI" from high-level principles and isolated tools toward a governance-aligned, evidence-centric reference architecture for real-world AI estates, including those running on hyperscale and HPC infrastructure (NIST, 2021, 2023; OECD, 2019; Jobin et al., 2019).

## 2 Background and Related Work

This section briefly situates PBSAI within three strands of work: AI for cybersecurity and multi-agent SOC assistance, systems security engineering and high-value asset protection, and AI governance and regulatory frameworks (Sommer and Paxson, 2010; NIST, 2021; NIST, 2023; Batool et al., 2025; Papagiannidis et al., 2025). We then present a short gap analysis motivating the need for a systems-level, multi-agent reference architecture for AI estates, followed by a discussion of PBSAI's positioning within existing security, SOC, and governance architectures.

### 2.1 AI for cybersecurity and multi-agent SOC assistance

AI has been applied to cybersecurity for more than a decade, most visibly in:

- log and alert analytics (for example, anomaly detection over SIEM data, clustering alerts to reduce noise);

- user and entity behavior analytics (UEBA), which apply statistical and machine learning models to identity and access patterns to detect account compromise or insider activity;

- endpoint and network protection, where models classify traffic, processes, or files as benign or malicious; and

- security orchestration, automation, and response (SOAR), which automate parts of incident response playbooks (Sommer and Paxson, 2010; Tounsi and Rais, 2018).

More recent work explores LLM-assisted SOC workflows and multi-agent defense, including the use of language models to summarize incident timelines, explain alerts, draft incident reports, assist analysts with query generation, and support human decision-making in complex cases, as well as early-stage experiments in multi-agent systems where agents specialize in triage, hunting, or reporting and coordinate via a shared context (Weidinger et al., 2022; Brundage et al., 2020). In practice, however, most deployed systems still look like:

- monolithic products that embed proprietary models inside a single tool (for example, an "AI-enhanced SIEM" or "EDR with AI"); or
- isolated automations built on top of existing SOAR platforms, often as custom scripts or playbooks.

These approaches can improve specific tasks, but they rarely provide a unified architectural pattern across tools, nor do they systematically integrate governance requirements, evidence collection, and cross-domain reasoning (Tounsi and Rais, 2018; Batool et al., 2025). The PBSAI ecosystem takes a different angle: it treats AI-for-cyber as a set of coordinated agent families spanning multiple domains, with shared context, schemas, and evidence semantics.

**2.2 Systems security engineering and high-value assets**

Systems security engineering work, particularly as captured in NIST SP 800-160 Volume 2, emphasizes that trustworthy systems result from structured engineering practices applied across the system lifecycle (NIST, 2021). SP 800-160v2 defines a range of techniques and approaches—such as Analytic Monitoring, Substantiated Integrity, Coordinated Defense, Diversity, Deception, Unpredictability, and Adaptive Response—and stresses that these must be incorporated into the design, implementation, and operation of complex systems rather than bolted on later.

Related practices for protecting high-value assets (HVAs), including those used by federal agencies and critical infrastructure operators, focus on:

- identifying Crown Jewels and critical functions;
- designing defense-in-depth architectures around them;
- integrating monitoring, response, and resilience measures; and
- conducting regular assessments to validate posture and detect drift (CISA, 2020; OMB, 2018).

In many environments, AI systems—particularly those that support key business processes or security operations—have effectively become HVAs. However, current guidance does not yet fully address AI estates as such, where models, orchestration, and hyperscale compute are first-class assets. PBSAI draws on systems security engineering principles and HVA practices but applies them to a multi-agent AI environment, encoding techniques like Analytic Monitoring and Substantiated Integrity into domain-specific agent responsibilities (NIST, 2021; CISA, 2020).

## 2.3 AI governance frameworks and regulatory developments

In parallel, a variety of AI governance frameworks and regulatory initiatives have emerged to address the risks of AI systems. The NIST AI Risk Management Framework (AI RMF) defines four functions—Govern, Map, Measure, Manage—to guide organizations in managing AI risk across the lifecycle, emphasizing governance structures, context and system mapping, risk measurement, and risk treatment (NIST, 2023). International and national bodies have proposed or adopted AI-specific standards and guidance, including ISO/IEC work on AI management and management systems (for example, ISO/IEC 42001) and security controls (for example, ISO/IEC 27001), as well as sectoral guidelines in areas such as health and finance (ISO/IEC, 2022, 2023).

Emerging AI regulations, exemplified by the European Union's AI Act, introduce obligations for "high-risk" AI systems, such as requirements for data governance, documentation, transparency, human oversight, robustness, cybersecurity, and post-deployment monitoring (European Union, 2024; OECD, 2019; Jobin et al., 2019). Broader ethics and governance discussions frame these obligations against high-level principles such as fairness, accountability, transparency, and human-centricity (Floridi and Cowls, 2019; Batool et al., 2025; Papagiannidis et al., 2025).

These frameworks and rules tend to be principle- and obligation-focused. They describe what must be true of AI systems (for example, that they be robust, transparent, and overseen by humans), but they do not prescribe how to:

- architect a multi-agent AI estate that satisfies these obligations;
- coordinate defensive AI agents and traditional controls under a common governance model; or
- capture and present evidence of compliance in a structured, machine-readable form (Brundage et al., 2020; NIST, 2023).

Organizations are left to bridge this gap themselves, often by layering policies on top of ad hoc technical solutions.

## 2.4 Gap analysis

Taken together, existing work and guidance leave several gaps that PBSAI aims to address.

**Model- and tool-centric focus vs. estate-level architecture.**
Much AI-for-cyber work focuses on specific models or tools—for example, UEBA, anomaly detection, or LLM triage assistants—rather than on the system-of-systems that makes up an AI estate (Sommer and Paxson, 2010; Tounsi and Rais, 2018). There is limited guidance on how to design a vendor-agnostic, multi-agent defensive architecture that spans governance, monitoring, response, resilience, and supply-chain domains.

**Principle-level governance vs. implementable patterns.**
AI governance and regulatory frameworks provide principles and high-level functions but lack implementable architectural patterns (OECD, 2019; NIST, 2023; European Union, 2024; Batool et al., 2025; Papagiannidis et al., 2025). Organizations lack a reference architecture that says, concretely, "these are the domains and roles you need, here is how agents interact, and here is how evidence flows to satisfy governance and regulatory expectations."

**Evidence and interoperability gaps.**
Most current implementations capture evidence in fragmented, tool-specific ways (for example, SIEM dashboards, SOAR logs, tickets), making it difficult to build a coherent, queryable evidence graph across the AI estate. There is no widely adopted pattern for AI-generated context envelopes, analytic schema contracts, or cryptographically signed outputs that preserve reasoning provenance across tools, vendors, and domains (Brundage et al., 2020).

**Limited integration of systems security engineering with AI governance.**
Techniques from systems security engineering—for example, Analytic Monitoring, Coordinated Defense, and Adaptive Response—are not systematically mapped into AI governance frameworks or multi-agent architectures (NIST, 2021, 2023). As a result, organizations struggle to demonstrate how their AI estates implement these techniques in practice, especially when leveraging hyperscale or HPC infrastructure.

The PBSAI Governance Ecosystem addresses these gaps by proposing a domain-structured, agent-based reference architecture for AI estates, grounded in systems security engineering, aligned with AI governance frameworks, and designed to produce the evidence needed for assurance and oversight (NIST, 2021, 2023; European Union, 2024; Batool et al., 2025; Papagiannidis et al., 2025).

## 2.5 Positioning within existing architectures

PBSAI does not propose yet another monolithic "AI product" or a replacement for existing SOC reference architectures. Instead, it sits above and around them as a management and evidence overlay for AI estates, intended to complement existing control catalogs, security architectures, and AI governance frameworks rather than duplicate them (NIST, 2020, 2021, 2023; ISO/IEC, 2022, 2023; European Union, 2024).

First, many environments are moving toward AI-enhanced point tools: SIEM platforms with embedded assistants, EDR products with proprietary models, or cloud security tools with built-in anomaly detection. These solutions can add value for specific tasks, but each uses its own internal models, schemas, and evidence formats, and is typically aligned with generic control catalogs rather than estate-level AI governance (NIST, 2020; Tounsi and Rais, 2018). Governance teams and SOC leaders are left with a mosaic of "AI inside X" capabilities that cannot easily be reasoned about as a whole estate. PBSAI assumes these tools exist and remain in place, but treats them as data sources and enforcement points under a shared set of domains, agents, MCP envelopes, and Output Contracts.

Second, pipeline-centric MLOps and model management architectures focus on the lifecycle of individual models—data collection, training, evaluation, deployment, and monitoring. They provide valuable abstractions for building and operating models, and are increasingly scoped by AI-specific management and assurance standards (ISO/IEC, 2023; NIST, 2023). However, they usually stop at the boundary of the model or service. Estate-level governance questions—such as how AI-assisted detections interact with identity systems, incident response, or resilience planning—are largely out of scope. PBSAI complements MLOps by treating models as one type of asset in Domain B (described in section 3.2) and by explicitly modeling how model-driven detections and decisions are mediated through domains such as C, D, G, H, and L, in line with AI governance expectations (OECD, 2019; Batool, et al., 2025; Papagiannidis, et al., 2025).

Third, traditional SOC reference architectures describe how to assemble SIEM, EDR, threat intelligence, SOAR, and case management into a coherent operational system. They are often grounded in control overlays and high-value asset protection guidance, such as federal HVA practices and sector-specific overlays (OMB, 2018; CISA, 2020). These designs typically treat AI, if at all, as either a future enhancement or a feature of individual tools. PBSAI can be viewed as a SOC-plus layer for AI estates: it retains the core SOC structure, but overlays a domain taxonomy, agent families, and evidence semantics that make AI-enabled behavior and governance explicit and traceable (NIST, 2021, 2023).

Finally, emerging "LLM-in-the-SOC" and multi-agent defense patterns often start from the AI side: they design a set of cooperating agents or assistants to help analysts with triage, hunting, or reporting. These efforts are motivated by broader work on AI risks and trustworthy AI development, especially for language models (Brundage et al., 2020; Weidinger et al., 2022; Jobin, et al., 2019). In practice, however, many current implementations are ad hoc, tied to a specific vendor or platform, and only loosely connected to formal governance or systems security engineering guidance (Floridi and Cowls, 2019; Papagiannidis, et al., 2025). PBSAI inverts this emphasis. It begins with domains, controls, and evidence requirements informed by systems security engineering and AI governance frameworks (NIST, 2021, 2023; ISO/IEC, 2022, 2023) and then defines multi-agent patterns that implement those responsibilities in a tool-agnostic way.

Table 1 summarizes these differences.

**Table 1 – PBSAI compared to existing architectural patterns**

| Pattern | Primary focus | Governance handling | Evidence model | Scale assumptions |
|---|---|---|---|---|
| AI-enhanced point tools | Individual products (SIEM, EDR) | Policy layered on top of each tool | Tool-specific logs and dashboards | Cloud / enterprise |
| Pipeline-centric MLOps | Model lifecycle | Mostly outside the model pipeline | Model-centric metrics and artifacts | Varies; usually per-service |
| Traditional SOC reference architectures | SOC tools and workflows | SOC processes and playbooks (often HVA-driven) | Case records, SIEM data, SOAR logs | Enterprise / large organization |
| LLM-in-the-SOC agent patterns | AI assistants for analysts | Often informal or tool-specific | Mixed; prompts, chat logs, tickets | Pilot to large, tool-dependent |
| PBSAI Governance Ecosystem | Estate-level domains and agents | Governance encoded in domains and envelopes | Structured Output Contracts, evidence graph | Enterprise → hyperscale / HPC-aware |

PBSAI is therefore best understood as a reference architecture for AI estates that can sit alongside existing SOC and MLOps designs. It assumes organizations will continue to use commercial tools and model platforms, but aims to give them a domain-structured, governance-aligned way to organize AI-enabled defenses and the evidence those defenses produce (NIST, 2021, 2023; European Union, 2024; Papagiannidis, et al., 2025).

## 3 Architecture Overview: 12-Domain Ecosystem and Minimal Stack

This section defines the overall structure of the PBSAI Governance Ecosystem. It first sets out the design goals and threat model for a tool-agnostic, high-assurance, AI-aware overlay on existing security stacks. It then introduces the twelve-domain taxonomy that organizes governance and operational responsibilities across the AI estate, describes the minimal baseline capabilities PBSAI assumes, and explains how automation is tied to systems security engineering techniques such as Analytic Monitoring, Substantiated Integrity, Coordinated Defense, and Adaptive Response (NIST, 2020, 2021, 2023; CISA, 2020; OMB, 2018).

Figure 1 presents a layered view of the PBSAI Governance Ecosystem. Governance and oversight (Domain A) sit above operational security domains (Domains B–K), supported by Domain L for security program enablement and AI validation. These domains are coordinated through a common agent pattern and MCP-style context envelopes that mediate between tools and policies. Underneath, existing tools, models, and infrastructure—including cloud and HPC environments—provide execution substrates. This layered structure makes explicit the separation of concerns that allows PBSAI to remain scale-invariant while preserving traceability, provenance, and policy alignment across deployment contexts.

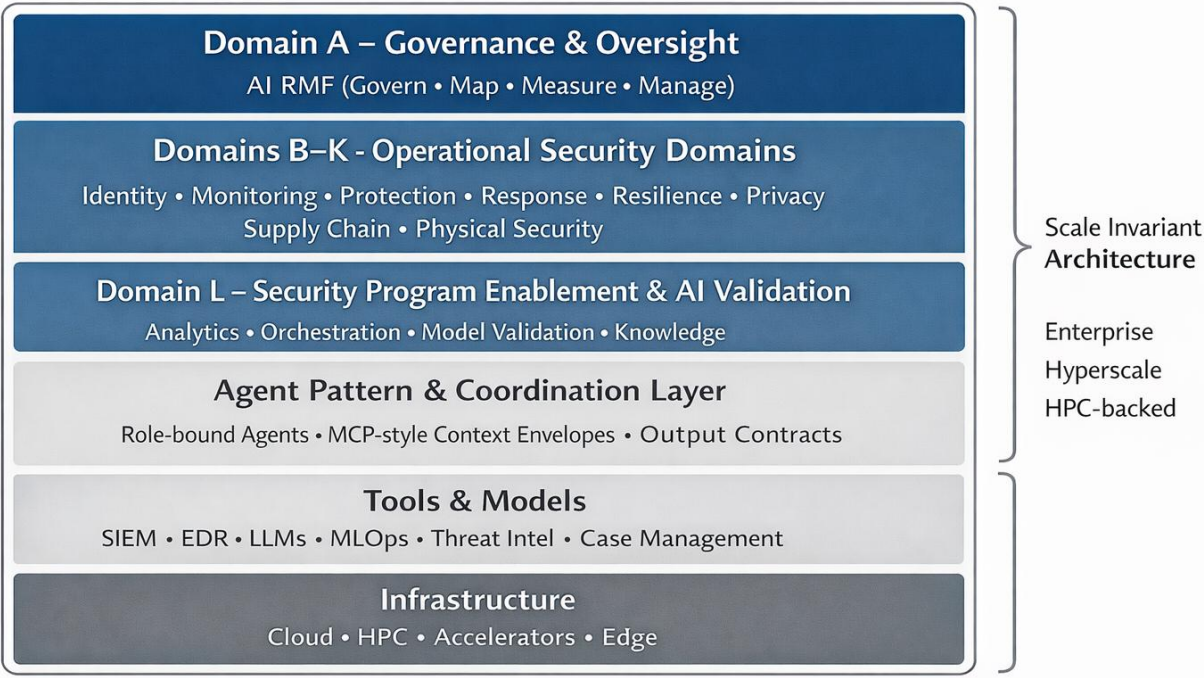

**Figure 1 – PBSAI Layered Architecture Overview.**
The PBSAI Governance Ecosystem organizes governance, operational domains, agent coordination, tooling, and infrastructure into layered responsibilities. The architecture separates policy from execution while preserving traceability and governance invariants across enterprise and hyperscale deployments.

### 3.1 Design goals and threat model

PBSAI is designed as a tool-agnostic management overlay for existing security and infrastructure stacks. Most organizations already have substantial investments in SIEM, EDR, IAM, cloud security, SOAR, and data protection tools. The ecosystem does not assume these are replaced wholesale. Instead:

- in many areas, agents fill gaps that commercial off-the-shelf (COTS) products do not address (for example, cross-tool evidence aggregation, policy-aware orchestration, AI validation);

- in some functions, implementers may choose to replace specific capabilities of a COTS product with agent-based functionality when there is a governance or interoperability advantage;

- core tools (for example, SIEM, EDR, IAM) remain the primary enforcement and sensing planes, while PBSAI provides the management, coordination, and evidence layer above them (Sommer and Paxson, 2010; Tounsi and Rais, 2018).

The architecture targets high-assurance and zero trust–oriented environments, consistent with modern systems security engineering and control guidance (NIST, 2020, 2021; ISO/IEC, 2022). It assumes:

- strong identity and access management for users, services, and agents;
- segmentation and least privilege across networks and workloads;
- explicit treatment of Crown Jewels and high-value assets; and
- continuous monitoring and validation of the security posture (CISA, 2020; OMB, 2018; NIST, 2021).

PBSAI is AI-aware but not AI-exclusive. It is built to coordinate:

- LLMs and tool-using agents;
- classical ML models (for example, UEBA, anomaly detection);
- rules-based and deterministic logic; and
- human workflows (Sommer and Paxson, 2010; Weidinger et al., 2022).

Agents can operate over non-AI telemetry and controls (for example, configuration baselines, network flows, physical access logs) while using AI selectively for classification, summarization, and reasoning where it adds value. The ecosystem is intended to work even if some domains have little or no AI adoption; in such cases, domains may be primarily deterministic and still benefit from the shared context and evidence model.

The blueprint is HPC- and hyperscale-aware. It anticipates environments where:

- streaming analytics and UEBA run on large clusters and accelerator pools;
- offline hunts and replay campaigns scan petabytes of telemetry;
- model training and validation for defensive AI models occurs on the same HPC or cloud accelerator infrastructure as production AI workloads (Weidinger et al., 2022; NIST, 2023).

PBSAI enforces a separation between management logic (agents, envelopes, contracts) and compute infrastructure, preserving architectural invariance across deployment environments.

The threat model includes both AI-specific and "classic" cyber threats (NIST, 2021; Weidinger et al., 2022):

- **AI-specific threats:** model misuse and abuse, prompt injection and tool-chain manipulation, data exfiltration via agents, poisoning of training or feedback data, autonomous loops that magnify errors or enable policy circumvention, and inappropriate use of LLMs in high-impact decisions.

- **Classic threats:** credential theft, privilege escalation, lateral movement, software and hardware supply-chain compromise, operational technology (OT) and industrial control systems (ICS) disruption, physical and environmental incidents, and insider threats (Sommer and Paxson, 2010; Tounsi and Rais, 2018; CISA, 2020).

- **Long-horizon cryptographic threats:** "harvest-now, decrypt-later" attacks against current cryptography and the need to plan for post-quantum (PQ) migration. PBSAI acknowledges these and defers detailed treatment to a deeper cryptography- and quantum-focused layer, but expects domains to be able to track which assets and data are at risk and which controls are PQ-ready (NIST, 2021; European Union, 2024).

To make the threat model more concrete, Table 2 sketches how representative threat categories map to PBSAI domains and to the systems security engineering techniques emphasized in NIST SP 800-160v2 (for example, Analytic Monitoring, Substantiated Integrity, Coordinated Defense, Adaptive Response) and to AI risk considerations in the NIST AI RMF (NIST, 2021; NIST, 2023).

**Table 2 – Representative threat categories and primary PBSAI coverage**

| Threat category | Example risks | Primary PBSAI domains | Key techniques / approaches |
|---|---|---|---|
| AI-specific misuse and abuse | Prompt injection, tool-chain manipulation, unsafe agent loops, overreach | D (Monitoring), F (Data), G (IR), L (AI validation) | Analytic Monitoring, Coordinated Defense, Adaptive Response |
| Data exfiltration and privacy harm | Exfiltration via agents, improper data access, privacy violations | C (Identity, Credential, and Access Management (ICAM)), F (Data Security), A (GRC) | Substantiated Integrity, Analytic Monitoring, Governance controls |
| Classic identity and infrastructure | Credential theft, lateral movement, | B (Asset/Config), C (ICAM), E (Protective), G (IR) | Coordinated Defense, |

| Threat category | Example risks | Primary PBSAI domains | Key techniques / approaches |
|---|---|---|---|
| | privilege escalation, service compromise | | Substantiated Integrity, Diversity |
| Supply chain compromise | Malicious or vulnerable software, models, or services | K (Supply Chain), B (Asset), I (Architecture), A (GRC) | Substantiated Integrity, Coordinated Defense |
| Long-horizon cryptographic threats | Harvest-now-decrypt-later, delayed compromise of stored data | A (GRC), F (Data), I (Architecture), K (Supply Chain) | Analytic Monitoring, Adaptive Response, lifecycle governance |

These categories are not exhaustive, but they illustrate how PBSAI treats threats as estate-level concerns: each category is addressed by coordinated domain responsibilities and agent behaviors rather than by isolated controls.

Out of scope for this paper are detailed operational runbooks and vendor-specific configurations. We do not attempt to specify a complete SOC operations manual, a full Model Context Protocol or equivalent standard, user interface designs for analyst tooling, or product-specific deployment recipes. The focus of PBSAI in this work is architectural: domains, agent roles, context envelopes, evidence contracts, and baseline assumptions that together define a governed, multi-agent defensive ecosystem for AI estates.

### 3.2 Domain taxonomy overview

The PBSAI ecosystem is organized into twelve domains that "ring" the AI estate, as shown in Figure 1. Each domain hosts agent families with specific responsibilities; domains interconnect via shared schemas and MCP-style context (NIST, 2021; NIST, 2023). Agents within each domain are enumerated using the convention [Domain Letter][Agent Number] (e.g., L3 denotes the SOAR Orchestration Agent in Domain L). A complete agent index is provided in Appendix A.

- **Domain A: Governance, Risk, and Compliance (GRC and Oversight).**
  Domain A defines policies, risk appetite, and control objectives for the AI estate. Its agents produce PolicyBundles, RiskAssessments, ComplianceStatus, and Metrics, consuming evidence from all other domains and driving AI RMF Govern and Manage functions (NIST, 2023; OECD, 2019; European Union, 2024).

- **Domain B: Asset, Configuration, and Change Management.**
  Domain B maintains inventories of assets, models, agents, data stores, and configurations, and tracks changes over time. Agents here detect drift, reconcile multiple data sources, and feed exposure and risk calculations in Domains D, E, and A (NIST, 2020, 2021).

- **Domain C: Identity, Credential, and Access Management (ICAM).**
  Domain C manages identities for humans, services, and agents, including authentication, authorization, and credential lifecycle. It provides the identity graph that UEBA, incident response, and governance rely on to answer "who did what, where, and with what authority" (NIST, 2020; ISO/IEC, 2022). A worked example of a Domain C agent that automates joiner–mover–leaver flows is given in Appendix B.

- **Domain D: Threat Intelligence, Situational Awareness, and Monitoring.**
  Domain D is the analytic sensing layer, ingesting logs, telemetry, and threat intelligence. Agents here perform correlation, anomaly detection, UEBA, and hunt activities, providing AlertClusters, AnomalyReports, BehavioralRisk scores, and ExposureMaps to other domains (Sommer and Paxson, 2010; Tounsi and Rais, 2018).

- **Domain E: Protective Technologies and Hardening.**
  Domain E focuses on preventive and detective controls: configuration baselines, patching, endpoint and network protections, and policy enforcement points. Agents recommend and sometimes orchestrate hardening actions, and evaluate control effectiveness over time (NIST, 2020).

- **Domain F: Data Security and Privacy.**
  Domain F manages data classification, encryption, data loss prevention (DLP) policies, and privacy impact assessments. It ensures that training and inference data for AI systems comply with privacy and lawful-basis requirements, and that data flows are understood and controlled (OECD, 2019; European Union, 2024; ISO/IEC, 2023).

- **Domain G: Incident Response and Digital Forensics (IR and DFIR).**
  Domain G orchestrates detection triage, case management, investigation, remediation, and post-incident analysis. Its agents create and manage IncidentCases, IncidentTimelines, RootCauseAnalyses, and LessonsLearned, tightly integrated with monitoring (D), identity (C), and governance (A) (NIST, 2020; CISA, 2020).

- **Domain H: Resilience, Continuity, and Recovery.**
  Domain H is responsible for backup strategies, disaster recovery, business continuity, and resilience exercises. Agents plan, execute, and evaluate disaster recovery (DR) tests and chaos experiments, generating evidence that supports continuity and resilience obligations (NIST, 2021; OMB, 2018).

- **Domain I: Security Architecture and Systems Engineering.**
  Domain I maintains architectural patterns, threat models, and design reviews for the AI estate. Agents check implementations against reference architectures, ensure that systems are aligned with security engineering principles and PBSAI patterns, and capture architectural decisions as evidence (NIST, 2021; Papagiannidis et al., 2025).

- **Domain J: Physical and Environmental Security.**
  Domain J correlates physical access, environmental sensors, and facility security controls. It provides context for incidents that span physical and cyber boundaries and ensures that physical risks to AI infrastructure (for example, data centers, network facilities) are monitored and managed.

- **Domain K: Supply Chain and Lifecycle Security.**
  Domain K tracks software, hardware, and service suppliers, along with advisories, vulnerabilities, and lifecycle states. Agents manage supplier risk, component provenance, and updates, and link these to assets and incidents in other domains (NIST, 2020; CISA, 2020).

- **Domain L: Security Program Enablement, Knowledge, and AI Validation.**
  Domain L hosts program-level enablement: security knowledge bases, analytics dashboards, SOAR orchestration, data curation for AI models, and independent AI validation. It coordinates cross-domain workflows and ensures that AI-for-cyber components are tested, monitored, and improved (Brundage et al., 2020; NIST, 2023).

Collectively, these domains form a defensive ring around the AI estate. Domains A, I, and L provide governance, architecture, and enablement; B, C, D, E, F, G, H, J, and K implement operational controls and analytics. Domains interconnect via shared contracts and MCP envelopes, enabling agents to exchange structured information without tight coupling to specific tools.

**3.3 Minimal secure AI stack and baseline assumptions**

PBSAI assumes a minimal secure AI stack is in place before deploying AI-enabled agents into critical security workflows. This baseline is not exhaustive, but it defines conditions under which the architecture can function as intended (NIST, 2020, 2021; ISO/IEC, 2022, 2023).

**Technical prerequisites.**
At the infrastructure and platform level, the ecosystem assumes:

- **Identity and ICAM foundation (Domain C).**
  A reliable identity platform for humans, services, and agents, including strong authentication, manageable authorization policies, and auditable identity events (NIST, 2020; ISO/IEC, 2022).

- **Centralized logging and telemetry (Domains B, D, F, J).**
  Telemetry from endpoints, network devices, cloud control planes, identity systems, physical sensors, and key applications is collected and retained in one or more analyzable stores. Time synchronization and basic normalization are in place (Sommer and Paxson, 2010; Tounsi and Rais, 2018).

- **Platform and agent attestation.**
  Mechanisms exist to attest to the integrity of infrastructure and workloads (for example, hardware roots of trust, measured boot, signed containers or binaries). Agents can be uniquely identified and their code provenance verified (NIST, 2020, 2021).

- **Segmentation and zero trust network posture (Domain E).**
  Networks and services are segmented according to risk; access is mediated by strong identity and policy; lateral movement is deliberately constrained. High-value assets and AI-critical components are placed behind additional controls (CISA, 2020; OMB, 2018; NIST, 2020).

- **Backup and continuity posture (Domain H).**
  Backups exist for critical data and systems; basic recovery paths are defined and periodically tested; continuity planning includes AI infrastructure and key AI-assisted defense components (NIST, 2021; OMB, 2018).

**Governance and policy prerequisites.**
At the governance and program level, PBSAI assumes:

- **Policy and control library.**
  A library of policies and controls exists, mapped at least coarsely to relevant frameworks (for example, NIST SP 800-53, SP 800-160v2 techniques, AI RMF functions, sectoral regulations) (NIST, 2020, 2021, 2023; ISO/IEC, 2022, 2023; European Union, 2024). This library can be referenced from MCP envelopes (policy_refs) and used by agents when evaluating compliance and risk.

- **Defined Human-in-the-Loop (HITL) thresholds and Crown Jewel inventories.**
  The organization has identified its high-value assets and critical business functions, and has defined when human approval is required for decisions affecting them. These thresholds are codified as constraints in MCP contexts (for example, actions requiring policy_cosign from specified roles) (CISA, 2020; OMB, 2018).

- **Evidence and schema registries.**
  There is a place to store structured evidence objects (for example, EvidenceManifests, IncidentTimelines, ModelValidationReports) and a registry of schemas (Output Contracts) that define their shape and validation rules. This allows agents to produce and consume evidence consistently and to support verifiable claims about system behavior (Brundage et al., 2020).

**HPC and hyperscale considerations.**

For organizations operating HPC clusters or large cloud estates, additional baselines are required:

- consistent attestation across multi-cluster and multi-cloud environments, so that agents can trust telemetry and workload identity regardless of where they run;

- unified identity and telemetry integration across on-prem HPC, cloud, and edge environments, avoiding "shadow" identity or logging domains that break end-to-end visibility;

- capacity planning and prioritization for AI-assisted defense workloads (for example, hunts, model validation, large-scale simulations), so that security jobs can run without starving critical business workloads and vice versa (NIST, 2023; Weidinger et al., 2022).

Where these baseline assumptions are not fully met, PBSAI can still be deployed in a degraded or partial mode, but its assurances and coverage are reduced. In practice, the baseline doubles as a readiness checklist for organizations considering a multi-agent AI defense architecture.

### 3.4 Automation and Alignment with NIST SP 800-160v2

For each domain and major agent family, PBSAI includes an internal analysis of automation feasibility and a mapping to NIST SP 800-160v2 techniques and approaches. This analysis asks (NIST, 2021):

- what tasks can be fully automated, which require HITL, and which should remain manual;

- how agents implement or support techniques such as Analytic Monitoring, Substantiated Integrity, Coordinated Defense, Diversity, Deception, Unpredictability, and Adaptive Response; and

- how these techniques are evidenced (for example, through specific Output Contracts or metrics).

As illustrative examples:

- In **Domain A**, Governance Policy and Risk Analysis agents implement Analytic Monitoring (over metrics and control health) and Coordinated Defense (propagating policy across domains), and drive Adaptive Response when they recommend changes to thresholds, controls, or AI model usage based on observed evidence.

- In **Domain B**, Asset Inventory and Config Drift agents implement Substantiated Integrity (comparing observed state to baselines with attestations) and Analytic Monitoring (continuous validation of asset and configuration inventories), providing foundational data for many other domains.

Other domains follow similar patterns: Domain D emphasizes Analytic Monitoring and Deception (through honeypots); Domains G and H emphasize Coordinated Defense and Adaptive Response; Domain K emphasizes Substantiated Integrity and coordinated supply-chain management (NIST, 2020, 2021; CISA, 2020).

For brevity, detailed per-domain tables are omitted from the main text but are available from the author upon request. The key point is that PBSAI offers a structured way to realize SP 800-160v2 techniques using concrete agent families and evidence contracts, making systems security engineering guidance directly applicable to AI estates (NIST, 2021; NIST, 2023).

## 4 Agent Pattern and MCP-Style Envelope

PBSAI's behavior emerges from many small agents working together, not from one monolithic "AI brain." This section defines the generic agent pattern and the MCP-style context envelope that every agent uses, along with how deterministic logic, LLM calls, schemas, provenance, and HITL fit together (Brundage et al., 2020; NIST, 2021; Weidinger et al., 2022).

### 4.1 Agent abstraction: small software workers between tools and policies

In PBSAI, agents are small software workers that sit between tools and policies. Each agent:

- subscribes to one or more input event types (for example, alert clusters, config changes, identity anomalies);
- runs deterministic logic, optionally aided by a bounded LLM call;
- emits one or more output events that conform to agreed schemas and carry a governance-aware context envelope.

At a high level, an agent implements a function of the form:

- output_event = Agent(input_event, context_envelope)

The input event carries the data to be processed (for example, a SIEM alert batch, a vulnerability record, a DR test result), and the context envelope carries information about mission, policy, constraints, and provenance. The agent's job is to transform events while respecting constraints and producing structured, signed outputs that other agents and humans can rely on (Brundage et al., 2020).

A condensed, worked example specification for a Domain C Identity Provisioning Agent (C1), illustrating this pattern in detail, is provided in Appendix B.

Agents are designed to be:

- loosely coupled to underlying tools (they call application programming interfaces (APIs) or consume streams, but do not depend on internal vendor formats);

- strongly typed at the boundaries (inputs and outputs must match schemas); and
- policy-aware, because the context envelope tells them what they are allowed to do and what evidence they must produce (NIST, 2020, 2021; ISO/IEC, 2023).

This abstraction is consistent across domains: a Governance Policy Agent in Domain A, a UEBA Agent in Domain D, or an AI Validation Agent in Domain L all follow the same basic pattern.

### 4.2 Lightweight formal model and invariants

While PBSAI is intended primarily as a practical architecture, it is useful to state a simple abstract model for agents and the context in which they operate. This clarifies what the ecosystem assumes and what properties it is intended to enforce.

We model each agent type as a function that consumes a single input event and an MCP-style context envelope, and produces zero or more output events:

- Let $E\_in$ be the set of input events that match a given input schema.
- Let $E\_out$ be the set of output events that match a specific Output Contract.
- Let $C$ be the set of valid MCP context envelopes.

An agent type A is then a function:

- $A : E\_in \times C \rightarrow list(E\_out)$

Given an input event $e\_in$ in $E\_in$ and a context $c$ in $C$, the agent produces a finite list of output events:

- outputs = $A(e\_in, c)$

Each output event must conform to its declared schema and be cryptographically bound to the agent that produced it. We can express this as:

- for all $e\_out$ in outputs:
  - validate_schema($e\_out$) = true, and
  - verify_signature($e\_out$) = true.

The MCP-style context envelope c is a structured record:

- c = (mission_id, thread_id, task_id, role, intent, policy_refs, constraints, decision_basis, provenance, classification, legal_hold)

where decision_basis includes references to evidence objects (for example, previous events, models, or reports) and confidence scores, and provenance encodes the agent's identity and attestation information.

On top of this abstract model, PBSAI aspires to maintain several ecosystem-level invariants. Informally:

1. **Traceability invariant.**
   For any security-relevant action event (for example, a containment action or privilege change) emitted by an agent, there must exist at least one associated RiskAssessment (or equivalent decision artifact) and at least one applicable PolicyBundle such that:

   - the action's context envelope includes policy_refs pointing to those policies; and
   - the action's decision_basis.evidence_refs includes the identifiers of the evidence objects that justified the action.

2. **Human-in-the-loop invariant.**
   For any action that affects a Crown Jewel asset or high-impact business function, the context envelope's constraints must require at least one human cosignature before execution (for example, a policy_cosign constraint referencing specific roles). Agents are not permitted to emit or execute such actions without the corresponding approvals being recorded in updated envelopes.

3. **Provenance and integrity invariant.**
   PBSAI is designed so that every event that changes system state (for example, configuration changes, access control changes, or containment actions) is produced by an agent whose provenance can be verified. In practice this means:

   - the event's signature validates against a known agent identity; and
   - the agent identity is bound to a current attestation for the agent's code and deployment pipeline.

These invariants are not formally proved in this paper, but they act as design constraints for the ecosystem. When implementers design agents, choose Output Contracts, or wire MCP envelopes into orchestration, they can treat these invariants as checks: if an action, decision, or state change cannot be traced back to policies, evidence, and a verified agent identity, then it does not yet meet PBSAI's expectations for a governed multi-agent architecture, in line with systems security engineering and assurance-oriented design practice (NIST, 2021; NIST, 2023).

### 4.3 MCP-style context envelope

Every agent invocation is wrapped in a context envelope inspired by Model Context Protocol (MCP) concepts. The envelope travels with the input, is updated by the agent, and is attached to the outputs. A typical envelope includes:

- **mission_id:** identifies the higher-level mission or campaign (for example, "protect crown jewels," "investigate incident X").

- **thread_id:** groups related tasks in a logical flow (for example, all steps in a specific incident).

- **task_id:** unique identifier for this particular agent task.

- **role:** the logical role the agent is playing (for example, "triage_assistant," "risk_analyst," "ai_validator").

- **intent:** short description of what the agent is trying to do (for example, "summarize_alerts," "prioritize_patches").

- **policy_refs:** references to policies and control sets that apply to this task (for example, "IR policy v3," "AI use policy v1," "data governance policy v2").

- **constraints:** machine-readable constraints (for example, "no direct containment," "HITL required for Crown Jewel actions," "read-only mode").

- **decision_basis:** structure capturing evidence_refs (ids of evidence objects used), confidence scores, and explanation_ref (pointer to a human-readable explanation or LLM-generated rationale).

- **provenance:** the agent's spiffe-style (cryptographically verifiable workload) identity, signing key id, and attestation information (for example, software bill of materials, build version).

- **classification:** security classification or sensitivity of the context (for example, "internal," "confidential," "regulated-data-present").

- **legal_hold:** flag or metadata indicating that data or outputs associated with this context are under legal or regulatory hold.

This envelope enables traceability and auditability: when a risk assessment or containment decision is questioned, reviewers can examine the mission, thread, policies, constraints, evidence, and agent identity associated with it (Brundage et al., 2020; NIST, 2021). The envelope also supports policy routing: orchestration agents in Domain L can route tasks to specific agents or require specific approvals based on mission, policy_refs, and constraints (NIST, 2023).

Legal and classification fields ensure that compliance and privacy constraints are visible at the point of action. For example, an agent might be allowed to analyze pseudonymized telemetry for one mission but must avoid accessing raw personal data for another. The envelope allows these distinctions to be encoded and enforced consistently across domains, aligning with regulatory and governance expectations for high-risk AI systems and regulated data (OECD, 2019; European Union, 2024; ISO/IEC, 2023; Floridi and Cowls, 2019).

## 4.4 Deterministic logic first, LLM assistance second

PBSAI adopts a deterministic-first, LLM-second pattern for agent behavior:

1. Agents apply deterministic logic wherever possible. This includes rule-based checks, threshold comparisons, graph traversals, and fixed formulas for risk or priority.

2. Only when deterministic logic cannot fully resolve a decision, or when a natural language transformation is genuinely helpful (for example, summarizing evidence), does the agent invoke an LLM.

3. The LLM call is bounded: the agent controls the prompt, constrains inputs to what is necessary, and post-processes outputs against schemas and constraints.

A simple example is risk scoring. An agent might compute:

- risk_score = (w1 * severity) + (w2 * confidence) + (w3 * asset_criticality)

where severity comes from the detection system, confidence from model or rule confidence, and asset_criticality from Domain B's inventory. The weights w1, w2, w3 are tuned using historical data and governance input. This risk_score is deterministic and reproducible. If necessary, an LLM might later be used to generate a human-readable explanation of why a particular alert or incident has a certain risk score, but the underlying decision logic remains deterministic, supporting governance expectations for traceability, auditability, and lifecycle oversight (NIST, 2023; Papagiannidis et al., 2025).

Bounded LLM calls are always followed by post-checks:

- outputs are validated against expected schemas;
- numerical values or classifications are checked for consistency with policy (for example, severity cannot exceed a maximum; certain actions require explicit constraints); and
- in high-risk contexts, HITL review may be required even if the LLM output passes schema checks (Weidinger et al., 2022; Batool et al., 2025).

This pattern ensures that LLMs supplement but do not replace deterministic, testable logic in security-critical workflows, and that model-assisted behavior remains governable and auditable (Brundage et al., 2020; NIST, 2021, 2023).

## 4.5 Schema contracts, provenance, and human-in-the-loop

PBSAI relies on Output Contracts (OCs) to standardize agent outputs. An OC is a schema for a particular kind of evidence or decision, such as:

- RiskAssessment
- ComplianceGap

- EvidenceManifest
- IncidentTimeline
- HardeningRecommendation
- ModelValidationReport

Agents must emit outputs that conform to these schemas. Before emission, outputs are:

- validated against the schema;
- assigned an idempotency key to prevent duplicate processing; and
- signed using the agent's key, with provenance information recorded in the envelope (Brundage et al., 2020; NIST, 2020, 2021).

A schema registry holds the definitions for all Output Contracts. This registry allows:

- agents to discover which schemas they can produce or consume;
- orchestrators to validate outputs before routing; and
- external tools or auditors to understand the structure of evidence (NIST, 2023).

Agent software itself is subject to supply-chain attestations. For example, an agent might carry a supply chain levels for software artifacts (SLSA) style attestation describing its build process, source repository, dependencies, and signing keys. The provenance field in the envelope can then reference this attestation, giving reviewers a way to verify that decisions are coming from agents built and deployed via trusted pipelines (NIST, 2021; ISO/IEC, 2022; CISA, 2020).

Human-in-the-loop (HITL) is modeled as explicit state and constraints, not as an informal notion. Constraints in the envelope can include:

- **policy_cosign:** list of roles or individuals who must approve an action before it proceeds;
- **HITL_on_gap_detected:** flag indicating that if certain conditions hold (for example, evidence gaps, conflicting signals), the agent must escalate rather than act automatically;
- **auto_open_ticket:** indication that an action should always result in a ticket or task for a human, even if the agent could proceed autonomously.

These constraints are interpreted by orchestrators and downstream agents. For example, if a containment agent receives a context with policy_cosign set to "IR.Manager, SystemOwner.ServiceX," it can prepare a proposed action but is not permitted to execute it until those approvals have been recorded and reflected in an updated envelope (NIST, 2023; European Union, 2024).

## 4.6 Lifecycle and deployment

Agents follow a full software lifecycle consistent with secure development practices, but tuned to the multi-agent context (NIST, 2020; ISO/IEC, 2022):

- development and testing in local or isolated environments;
- integration into a CI/CD pipeline that builds, tests, signs, and packages agents;
- deployment to a canary environment, where they process real or replayed data under close observation;
- gradual promotion to production, with clear rollback paths.

During this lifecycle, the ecosystem tracks service level objectives (SLOs) for agents, such as:

- **Ack_p95:** "95 percent of tasks acknowledged within X seconds";
- **false_alert_rate:** "fraction of agent-generated alerts or recommendations later judged unnecessary or incorrect";
- **coverage:** "fraction of applicable events or assets that the agent actually processes or evaluates."

These SLOs allow governance and engineering teams to evaluate whether agents are performing as intended and whether automation levels should be adjusted up or down (NIST, 2021, 2023).

Rollback criteria are defined ahead of time. For example, an agent or model may be rolled back if:

- its false_alert_rate exceeds a threshold over a defined period;
- its Ack_p95 or similar timeliness metric degrades beyond acceptable bounds;
- it generates outputs that repeatedly conflict with policy or require frequent human overrides; or
- its underlying attestation or dependencies can no longer be trusted (for example, due to a supply-chain advisory) (CISA, 2020; OMB, 2018; NIST, 2021).

Because agents operate under MCP-style envelopes and produce signed outputs with clear provenance, rolling back an agent does not destroy evidence of its past behavior. Historical decisions and outputs remain traceable, and new agents or versions can be evaluated against the same contracts and SLOs. This makes the ecosystem amenable to iterative improvement and experimentation while preserving accountability (Brundage et al., 2020; NIST, 2021, 2023).

## 5 Mapping to NIST AI RMF and NIST SP 800-160v2

This section connects the PBSAI ecosystem to existing guidance on systems security engineering and AI risk management. It sketches how key NIST SP 800-160v2 techniques—such as Analytic Monitoring, Substantiated Integrity, Coordinated Defense, and Adaptive Response—are realized through domain responsibilities and agent behaviors, and then maps PBSAI's domains and agent families to the NIST AI Risk Management Framework functions (Govern, Map, Measure, Manage), highlighting how the architecture supports regulatory objectives for high-risk AI systems and sectoral regimes (NIST, 2020, 2021, 2023; CISA, 2020; OMB, 2018).

### 5.1 NIST SP 800-160v2 techniques in the ecosystem

NIST SP 800-160v2 emphasizes that trustworthy systems are engineered by applying a set of techniques and approaches consistently across the lifecycle (NIST, 2021). PBSAI encodes several of these directly into domain responsibilities and agent behaviors. Four techniques are particularly central: Analytic Monitoring, Substantiated Integrity, Coordinated Defense, and Adaptive Response.

**Analytic Monitoring.**
Analytic Monitoring calls for continuous analysis of system behavior, not just raw logging (NIST, 2021). In PBSAI, this is primarily realized in:

- **Domain D (Threat Intelligence, Situational Awareness, and Monitoring)**, where agents such as Threat Intel Aggregator, SIEM Analyst, UEBA Analysis, and Autonomous Hunter perform correlation, anomaly detection, and hunt operations over telemetry. They generate structured outputs like AlertClusters, AnomalyReports, BehavioralRisk scores, and HuntFindings (Sommer and Paxson, 2010; Tounsi and Rais, 2018).

- **Domain L (Security Program Enablement and AI Validation)**, where Security Analytics and AI Validation agents analyze agent performance, model behavior, and coverage metrics. They produce ModelValidationReports, QualityMetrics, and SystemHealth indicators (Brundage et al., 2020; NIST, 2023).

Together, Domains D and L ensure that monitoring is analytic and continuous, not just a collection of logs and dashboards: they turn telemetry into actionable, evidence-backed understanding of the AI estate (NIST, 2021, 2023).

**Substantiated Integrity.**
Substantiated Integrity requires that the system can justify claims about its state with credible evidence (NIST, 2021). PBSAI implements this through:

- **Domain B (Asset, Configuration, and Change)**, where Asset Inventory and Config Drift agents reconcile inventories from multiple sources, compare observed state to baselines, and rely on platform attestations for critical components. Their outputs (AssetInventory, ConfigDrift, IntegrityCheck) form the basis for answering "what is actually running, where, and in what state?" (NIST, 2020, 2021).

- **Domain K (Supply Chain and Lifecycle Security)**, where Supplier Risk and Component Provenance agents track software, hardware, and service dependencies, advisories, and lifecycle states. They record provenance, SLSA-style attestations, and vulnerability status for components (CISA, 2020; NIST, 2020).

By linking B and K through shared contracts and evidence manifests, PBSAI provides substantiated integrity both for running systems and for the components and suppliers they depend on (NIST, 2021; CISA, 2020).

**Coordinated Defense.**
Coordinated Defense stresses that security measures must reinforce each other across the system, rather than operate as isolated controls (NIST, 2021). In PBSAI, coordinated defense is driven by:

- **Domain A (Governance, Risk, and Compliance)**, which defines policies, risk appetite, and control objectives and receives evidence from all other domains (NIST, 2023; OECD, 2019; European Union, 2024).

- **Domain L (Program Enablement and Orchestration)**, particularly L3 SOAR Orchestration and L1 Security Knowledge agents, which coordinate workflows across B, C, D, E, F, G, H, J, and K (Tounsi and Rais, 2018; Brundage et al., 2020).

- **Domain I (Security Architecture and Systems Engineering)**, which maintains the reference architectures and ensures that new systems and changes align with PBSAI patterns and defense-in-depth principles (NIST, 2021; Papagiannidis et al., 2025).

These domains encode how policies, architecture, and orchestrated workflows link preventive, detective, and reactive controls together, so that improvements or failures in one area are visible and compensated for in others (NIST, 2021, 2023).

**Adaptive Response.**
Adaptive Response focuses on the system's ability to adjust defenses based on changing conditions and evidence (NIST, 2021). PBSAI supports this through:

- **Domain G (Incident Response and DFIR)** and **Domain H (Resilience, Continuity, and Recovery)**, where Case Manager, Remediation Planner, DR Orchestrator, and Chaos/Exercise agents recommend or enact changes to configurations, playbooks, and continuity plans based on incident and resilience findings (NIST, 2020; CISA, 2020; OMB, 2018).

- **Domain D (Monitoring)**, where detection and UEBA models can be tuned based on new threat intelligence and validation results from Domain L (Sommer and Paxson, 2010; Weidinger et al., 2022; NIST, 2023).

Adaptive behavior is constrained by governance: policy changes and automation level adjustments are surfaced as RiskAssessments or PolicyBundle updates in Domain A, and encoded as new constraints or thresholds in MCP envelopes, ensuring that adaptation remains evidence-based and governed, not ad hoc (NIST, 2021, 2023; Batool et al., 2025).

### 5.2 Mapping to NIST AI RMF functions

The NIST AI RMF defines four functions—Govern, Map, Measure, Manage—that organizations should implement across their AI lifecycle (NIST, 2023). PBSAI's domains and agent families align with these functions as shown in Table 3.

**Table 3 – PBSAI domains mapped to NIST AI RMF functions**

| AI RMF Function | Primary PBSAI Domains | Example Responsibilities |
| --- | --- | --- |
| Govern | A (GRC), I (Architecture), L | Policies, risk appetite, reference architectures, SOAR orchestration strategy, AI usage policies, model validation governance |
| Map | B (Asset/Config), C (ICAM), F (Data Security), K (Supply Chain) | AI system inventory, data flows and classifications, identity mapping, component and supplier tracking, risk context |
| Measure | D (Monitoring), L2 (Analytics), L5 (AI Validation) | Detection and UEBA metrics, telemetry analytics, model performance and robustness evaluation, coverage and false alert rates |
| Manage | E (Protective Controls), G (Incident Response / Digital Forensics Incident Response), H (Resilience), L3 (SOAR) | Hardening and patching, incident triage and remediation, DR and continuity operations, playbook execution and automation controls |

In practice, all domains contribute to multiple functions, but the mapping provides a clear starting point for organizations that want to show that their AI estate design covers AI RMF expectations (NIST, 2023). For example, an external assessor can trace how Govern artifacts (policies in Domain A) are implemented via Map (inventories and data flows in B, C, F, K), assessed via Measure (analytics and validation in D, L2, L5), and acted on via Manage (controls, incidents, and resilience in E, G, H, L3) (European Union, 2024; Papagiannidis et al., 2025).

## 5.3 Regulatory alignment (high-level)

Regulatory regimes for high-risk AI systems typically emphasize obligations such as:

- sound data governance and documentation of data lineage, quality, and representativeness;

- appropriate human oversight and clear roles and responsibilities;

- robustness and cybersecurity measures, including monitoring and incident handling; and

- comprehensive technical documentation and logs that can support audits and investigations (OECD, 2019; European Union, 2024; Floridi and Cowls, 2019; Batool et al., 2025).

PBSAI provides a structure for meeting these obligations:

- **Domain F** and associated agents handle data classification, access, and lawful-basis rules;

- **Domains A, C, and G** encode human oversight and decision roles via policy, identity mappings, and HITL constraints in MCP envelopes;

- **Domains D, E, G, and H** implement robustness and incident response patterns; and

- the evidence graph (fed by EvidenceManifests, IncidentTimelines, ModelValidationReports, and related contracts) supplies audit-ready evidence for regulators and internal oversight (Brundage et al., 2020; NIST, 2021, 2023).

Sector-specific regulations—for example, in finance, healthcare, or critical infrastructure—also introduce requirements around incident reporting, continuity, and supply-chain assurance (CISA, 2020; OMB, 2018; ISO/IEC, 2022, 2023). PBSAI addresses these by:

- giving **Domains G and H** explicit responsibilities for incident classification, reporting artifacts, DR plans, and the evidence to show that continuity and recovery procedures are tested;

- assigning **Domain K** responsibility for supplier and component risk, including tracking advisories, vulnerabilities, and lifecycle states; and

- linking these domains to governance (**Domain A**) through structured outputs and MCP contexts, so that regulatory reporting and oversight are built on the same evidence used for day-to-day operations (NIST, 2020, 2021; European Union, 2024).

The application scenarios show how these mappings play out in concrete settings: a medium-scale enterprise SOC using PBSAI as an overlay on an existing tool stack, and a hyperscale/HPC-backed environment where PBSAI coordinates streaming analytics, replay campaigns, and AI validation. In both cases, the same domain and agent structure provides a direct line of sight from regulatory objectives and AI RMF functions to the architecture and behavior of the AI estate (NIST, 2023; Papagiannidis et al., 2025).

## 6 Application Scenarios

This section sketches two illustrative deployments of the PBSAI ecosystem and draws out cross-cutting lessons. The first scenario shows a medium-scale enterprise SOC where PBSAI is layered as a multi-agent overlay on top of an existing SIEM/EDR/IAM/SOAR stack, adding MCP envelopes and evidence contracts without replacing core tools. The second scenario considers a hyperscale/HPC defensive estate, where PBSAI coordinates streaming analytics, large-scale replay and hunting, data curation, and AI validation on accelerator-rich infrastructure. The section closes with common themes across both environments, including incremental adoption, integration friction around identity, telemetry, and schemas, the central role of human factors, and the way HPC changes scale rather than the underlying structure of the ecosystem. The goal is to show how the same domain and agent structure adapts to very different scales and levels of maturity (NIST, 2021, 2023; CISA, 2020).

### 6.1 Scenario 1 – Enterprise SOC overlay

Consider a medium-scale, cloud-first enterprise with:

- a single primary cloud provider plus some on-prem workloads;
- an existing SIEM aggregating logs from endpoints, network devices, and cloud control planes;
- EDR agents on laptops and servers;
- an identity provider with single sign-on (SSO) and multi-factor authentication (MFA);
- a basic SOAR platform automating a handful of playbooks; and
- a SOC team of roughly a dozen analysts and engineers.

The organization wants to use AI to improve triage and investigation and to strengthen governance, but does not want to rip and replace its current tools. PBSAI is introduced as a multi-agent overlay, starting with a minimal set of domains and agents: A (GRC), B (Asset/Config), D (Monitoring), G (IR/DFIR), and L (Program Enablement/SOAR) (Sommer and Paxson, 2010; Tounsi and Rais, 2018; NIST, 2023).

In Domain B, Asset and Config agents pull data from the configuration management database (CMDB), cloud inventory APIs, and SIEM asset tables, reconcile inconsistencies, and produce AssetInventory and ConfigDrift contracts. These are wrapped in MCP envelopes that capture mission_id (for example, "SOC-operations"), policy_refs (for example, "baseline hardening policy"), and constraints (for example, "read-only; no automated changes"). Domain D agents—Threat Intel Aggregator, SIEM Analyst, UEBA Analysis—consume alerts and logs from the SIEM, enrich them with threat intel, calculate BehavioralRisk scores, and output AlertCluster and AnomalyReport contracts under their own MCP contexts (Sommer and Paxson, 2010; Tounsi and Rais, 2018).

Domain G deploys a Case Manager Agent that listens for Domain D outputs. When an AlertCluster with risk_score above a policy-driven threshold arrives, the agent opens an incident, assigns a thread_id, and spawns a Timeline Reconstruction Agent to pull related events into an IncidentTimeline contract. This evidence is stored in the registry and linked to the incident via the envelope, supporting traceability and post-incident review (NIST, 2020; CISA, 2020).

Domain L integrates with the existing SOAR platform. L3 SOAR Orchestration subscribes to incidents from Domain G and uses MCP envelopes to decide which SOAR playbook to trigger. For example, a medium-severity identity-related incident might cause L3 to call an existing "suspicious login" playbook, passing along context (asset, identity, policy constraints). SOAR executes the usual steps—collect more logs, disable tokens, notify the user—while PBSAI agents add structured evidence and produce a final IncidentSummary contract (NIST, 2020; Tounsi and Rais, 2018).

Domain A Governance and Risk agents periodically consume IncidentSummary, RiskAssessment, and ControlStatus outputs from other domains. They roll these into higher-level RiskAssessment and Metrics contracts for leadership and auditors, answering questions like "Which Crown Jewels are repeatedly affected?" or "How often do we see delayed detection in specific business units?" This supports AI RMF Govern and Manage functions by tying SOC outcomes back to governance artifacts (NIST, 2023; OECD, 2019).

A concrete flow looks like:

- **D2: SIEM Analyst Agent** clusters a burst of login failures and unusual geolocation activity, creates an AlertCluster with risk_score, and references raw alerts.

- **G1: Incident Triage Agent** receives the cluster, opens an incident, and instructs **G4: Forensic Analysis Agent** to reconstruct an IncidentTimeline.

- **L3: SOAR Orchestration Agent** reads the MCP envelope, determines this is a medium-severity identity event with no Crown Jewels involved, and triggers an automated playbook to revoke tokens and force password resets.

- **A2: Risk Analysis Agent** (and, if metrics are updated, **A7: Metrics Aggregator Agent**) later ingests the IncidentSummary and EvidenceManifest, updating risk metrics and, if needed, recommending policy adjustments (e.g., stronger MFA requirements for a specific application).

Throughout this scenario, PBSAI does not replace SIEM, EDR, IAM, or SOAR. Instead, it adds MCP envelopes, agent logic, and evidence contracts that bring coherence, traceability, and governance semantics to workflows the SOC already runs (NIST, 2021, 2023; Brundage et al., 2020).

### 6.2 Scenario 2 – Hyperscale/HPC defensive estate

Now consider a large organization (a major cloud provider, financial institution, or critical infrastructure operator) with:

- multiple HPC clusters and GPU/accelerator pools;
- large-scale production AI workloads (for example, recommendation systems, forecasting models, optimization solvers) running on that infrastructure;
- a global, high-volume telemetry fabric spanning on-prem, cloud, and edge; and
- a mature security program that already uses machine learning for detection at scale (Sommer and Paxson, 2010; NIST, 2021).

The organization wants to turn its HPC capabilities into a hyperscale AI defensive estate while maintaining strict governance and human oversight. PBSAI is deployed across all domains, but Domains D and L, and their interplay with B, C, G, H, and K, are especially prominent (NIST, 2021, 2023).

In Domain D, streaming telemetry (logs, network flows, identity events, application traces) is ingested into a central analytics fabric running on HPC and cloud accelerators. A UEBA Analysis Agent trains and runs large behavioral models across millions of identities and entities. It outputs BehavioralRisk contracts, each capturing an identity, time window, risk_score, and explanatory features, under envelopes that include mission_id ("global-UEBA"), policy_refs ("UEBA use policy"), and classification ("internal-plus-sensitive"). An Autonomous Hunter Agent runs recurrent replay campaigns on historical telemetry, using new rules or models to search for missed campaigns; its HuntFinding and RetrospectiveIncident outputs feed Domain G and Domain A (Sommer and Paxson, 2010; Tounsi and Rais, 2018).

Domain L4 Data Curation Agents select training and validation data for defensive models from telemetry streams and storage. They ensure that privacy and lawful-basis constraints from Domain F are respected (for example, only pseudonymized fields used where required) and that samples are representative of diverse business units and geographies. Curated data sets are registered as TrainingDataset and EvaluationDataset contracts, with provenance and constraints encoded in MCP envelopes (OECD, 2019; European Union, 2024).

Domain L5 AI Validation Agents then orchestrate extensive evaluation of candidate models on HPC infrastructure. They run offline tests over curated data, scenario-based replays (for example, known ransomware campaigns, stealthy insider activity), and robustness checks. The results are packaged as ModelValidationReport contracts, detailing metrics (precision, recall, false positive rates), stability, known weaknesses, and recommended deployment constraints (for example, "advisory mode only," "HITL required for containment suggestions") (Weidinger et al., 2022; Batool et al., 2025). These reports are consumed by governance agents in Domain A for risk and policy decisions, and by L3 SOAR Orchestration Agent in Domain L for deployment configuration, aligning with the AI RMF Measure and Manage functions (NIST, 2023).

Automated containment is tightly governed using constraints and HITL. When a high-confidence BehavioralRisk or HuntFinding arrives:

- L3 SOAR Orchestration considers the model's validation status, the asset's criticality (from Domain B), and the associated policies (from Domain A).

- For low-impact assets (for example, non-critical user endpoints), the context may allow fully automated containment such as isolating a machine or revoking tokens.

- For high-impact assets (Crown Jewels, production clusters, critical services), the MCP envelope includes constraints such as policy_cosign = ["IR.Manager", "SystemOwner.ServiceX"] and HITL_on_gap_detected = true. In these cases, agents prepare proposed actions and supporting evidence, but execution requires explicit approval in Domain G and is recorded as part of the evidence graph (NIST, 2023; European Union, 2024).

The same HPC and accelerator infrastructure is thus used to:

- train and run large defensive analytics;

- replay and hunt across vast telemetry archives;

- validate and monitor defensive models at scale; and

- empower orchestrated, constrained automation that respects governance boundaries (NIST, 2021, 2023; Papagiannidis et al., 2025).

PBSAI's domains, agents, and MCP envelopes provide the structure that keeps this complexity governable.

### 6.3 Lessons

Several cross-cutting lessons emerge from these scenarios.

**Incremental adoption is essential.**
Organizations rarely have the appetite or capability to deploy a full multi-agent ecosystem in one step. The enterprise SOC overlay shows how PBSAI can start with a thin slice—a handful of agents in A, B, D, G, and L working in advisory or semi-automated mode—on top of existing tools. Over time, more domains and deeper automation can be added, guided by evidence and SLOs. The hyperscale scenario represents a later stage, not a starting point (NIST, 2021, 2023; CISA, 2020).

**Integration friction centers on identity, telemetry, and schemas.**
In both environments, the hardest problems are:

- identity integration (Domain C): unifying identities across directories, cloud accounts, and services so that agents can reason about "who did what";

- telemetry normalization (Domains B, D, F, J): bringing diverse log formats and event sources into analyzable, consistent streams; and

- schema definition/adoption: agreeing on Output Contracts and MCP envelope conventions so that agents and tools can interoperate (NIST, 2020; ISO/IEC, 2022, 2023).

Investment in these foundations is a prerequisite for PBSAI's full benefits and is often where early deployments uncover latent weaknesses.

**Human factors are as important as algorithms.**
HITL thresholds, approval flows, and operator trust determine whether automation is accepted or resisted. Analysts must see that:

- agents are not making irreversible changes without oversight;

- evidence and explanations are transparent and accessible;

- rollback paths exist for agents and models that misbehave.

Explicitly modeling HITL as constraints in MCP envelopes, surfacing rationales in human-friendly form, and involving SOC and GRC staff in threshold and constraint design are crucial for success (Floridi and Cowls, 2019; Weidinger et al., 2022; Batool et al., 2025).

**HPC changes scale, not structure.**
The hyperscale scenario demonstrates that HPC and accelerators allow more ambitious analytics—global UEBA, massive replay, large-scale validation—but they do not alter the fundamental structure of PBSAI. The same twelve domains, agent patterns, envelopes, and evidence contracts apply. Organizations without HPC can still adopt the architecture at smaller scale; those with HPC can exploit it to cover more data and more complex analyses, using the same governance and systems engineering scaffolding (NIST, 2021; Papagiannidis et al., 2025).

## 7 Discussion and Conclusion

This section reflects on the advantages and limits of the PBSAI Governance Ecosystem and outlines directions for future work. Section 7.1 discusses the main benefits and trade-offs of adopting a twelve-domain, agent-based architecture for AI estates. Section 7.2 considers PBSAI as an open ecosystem and notes key limitations of the current work. Section 7.3 presents testable hypotheses and an evaluation agenda, and Section 7.4 concludes the paper by summarizing the core contribution and future research directions.

## 7.1 Benefits and trade-offs

PBSAI's primary benefit is that it brings coherence to environments that are otherwise assembled from disconnected tools and point solutions. By organizing responsibilities into a twelve-domain taxonomy and expressing capabilities as agent families with shared MCP-style envelopes and Output Contracts, the blueprint turns a scattered set of products into a structured ecosystem (NIST, 2021, 2023). Governance, operations, and engineering can all refer to the same domains, roles, schemas, and evidence artifacts when they talk about how the AI estate is defended, aligning with evidence- and risk-centric approaches in contemporary AI and security frameworks (OECD, 2019; European Union, 2024). This structure supports evidence-centric governance: rather than relying on informal narratives and screenshots, organizations can build a queryable evidence graph that directly links policies, assets, controls, models, agents, incidents, and outcomes, echoing calls for auditable AI assurance and documentation in recent work (Floridi and Cowls, 2019; Weidinger et al., 2022). At the same time, the architecture increases cross-domain visibility—identity, data protection, incident response, resilience, and supply-chain risk can all be viewed through consistent schemas and envelopes—and provides a disciplined way to use AI-for-cyber, with bounded LLM assistance and explicit validation and oversight mechanisms (Brundage et al., 2020; Batool et al., 2025).

These benefits come with real trade-offs. The ecosystem introduces additional complexity on top of already complex SOC and infrastructure environments (Sommer and Paxson, 2010; Tounsi and Rais, 2018). Even in incremental deployments, organizations must define schemas, MCP envelope conventions, and agent responsibilities; set up evidence and schema registries; and adapt workflows to include new agent-generated artifacts (NIST, 2021). The architecture also adds operational overhead, particularly in the early stages, as teams tune SLOs, thresholds, and constraints and learn how to interpret and trust agent outputs (NIST, 2020, 2023). Finally, PBSAI is explicitly dependent on strong identity and attestation foundations: if identities for users, services, and agents are fragmented or untrustworthy, or if platform attestation is unavailable, some of the architecture's assurances are weakened (NIST, 2020; ISO/IEC, 2022). In those cases, PBSAI may surface gaps and inconsistencies that must be addressed before its full potential can be realized, consistent with systems security engineering guidance that emphasizes prerequisite foundations for trustworthy architectures (NIST, 2021).

## 7.2 Open ecosystem and limitations

PBSAI is intended as a blueprint, not a product. The agent patterns, MCP-style envelopes, and Output Contracts described here are meant to serve as starting points for a shared, open ecosystem, not proprietary interfaces or closed implementations. One possible evolution is the emergence of community- or institution-led initiatives, similar in spirit to existing open security initiatives, that curate and publish:

- an open catalog of Output Contracts and context envelope fields;
- reference domain responsibilities and example agent behaviors;
- interoperability profiles that vendors and open-source projects can align with (NIST, 2021; OECD, 2019).

In such a model, PBSAI becomes a reference architecture and shared vocabulary that multiple stakeholders can adopt and extend. Vendors could implement PBSAI-compatible agents or connectors; open-source projects could provide reusable building blocks for specific domains; and regulators or sector bodies could reference PBSAI-like patterns when describing expectations for AI-enabled defenses and evidence collection (European Union, 2024; NIST, 2023).

At the same time, the current work has important limitations. The architecture is conceptual and pattern-based; it is not yet backed by systematic empirical validation or industry-standard benchmarks. Open questions include how PBSAI-style deployments affect detection and response effectiveness, false alert rates, analyst workload, and governance outcomes in different organizational contexts (Sommer and Paxson, 2010; Papagiannidis et al., 2025). There is also a need for reference implementations and testbeds—modular, open implementations of agents, schemas, envelopes, and registries that practitioners can experiment with and researchers can use to study multi-agent defensive systems under controlled conditions (NIST, 2021; Weidinger et al., 2022; Batool et al., 2025). Without these, organizations must build and validate many components themselves, which raises the barrier to adoption and slows the feedback loop needed to refine the blueprint.

## 7.3 Hypotheses and evaluation agenda

Although this paper is primarily architectural, PBSAI is intended to be empirically testable. In particular, the blueprint suggests several hypotheses that can be evaluated through controlled experiments, pilot deployments, and comparative studies, in line with calls for evidence-based AI governance and risk management (NIST, 2023; Papagiannidis, Mikalef and Conboy, 2025):

- **H1 – Evidence-centric architecture improves auditability and governance clarity.**
  Organizations that adopt PBSAI's domain taxonomy, MCP-style envelopes, and Output Contracts will be able to answer audit and oversight questions (for example, "who did what, under which policies, based on what evidence?") more completely and with less manual effort than comparable organizations using ad hoc, tool-specific evidence practices.

- **H2 – Multi-agent overlays reduce investigation latency and improve coverage.**
  In SOC environments with similar tool stacks, a PBSAI-style multi-agent overlay (Domains A, B, D, G, L) will reduce mean time to triage and increase incident and exposure coverage, relative to baselines that rely solely on embedded "AI inside" features or traditional SOAR playbooks (Tounsi and Rais, 2018; NIST, 2023).

- **H3 – Bounded LLM use with HITL constraints reduces harmful automation errors.**
  Agent patterns that follow PBSAI's "deterministic first, LLM second" rule, combined with explicit HITL constraints in context envelopes, will exhibit lower rates of harmful or policy-violating automation outcomes than naive "LLM-in-the-loop" automations with similar model capabilities (Brundage et al., 2020; Weidinger et al., 2022).

- **H4 – HPC-backed deployments increase scale, not governance fragility.**
  When PBSAI is deployed on hyperscale or HPC infrastructure, increased analytic and validation capacity (for example, global UEBA, large-scale replay) will improve detection and validation coverage without degrading the traceability, provenance, and oversight invariants defined for smaller-scale deployments, provided that baseline identity and attestation assumptions hold (NIST, 2021, 2023).

- **H5 – Quantum-assisted analysis expands feasible evaluation scope in high-dimensional settings.**
  For large-scale AI estates with combinatorial attack surfaces or multi-objective governance constraints, quantum-assisted optimization or sampling methods may enable evaluation of larger state spaces or more complex dependency graphs than classical approaches under comparable time constraints.

Testing these hypotheses will require reference implementations, shared benchmarks, and longitudinal studies across different sectors and scales. Section 6 sketches qualitative scenarios; future work should turn these into quantitative evaluation programs that measure both operational metrics (for example, detection, response, false alert rates) and governance outcomes (for example, audit effort, quality of evidence, regulator-facing clarity), and, for high-dimensional risk and dependency analyses, comparative performance of classical and quantum-assisted optimization approaches under controlled governance and traceability constraints.

## 7.4 Conclusion

This paper has presented the PBSAI Governance Ecosystem as a multi-agent reference architecture for securing AI estates—socio-technical systems that span models, agents, data pipelines, human workflows, security tooling, and hyperscale infrastructure. Rather than treating AI governance, cyber defense, and AI-for-cyber capabilities as separate domains, PBSAI integrates them within a single systems security engineering framework grounded in domain-structured responsibilities, reusable agent patterns, MCP-style context envelopes, and structured Output Contracts.

The central claim of PBSAI is architectural: that governance-aligned AI defense requires explicit coordination across domains, verifiable provenance for agent actions, and an evidence-centric model in which policies, telemetry, models, and operational decisions are linked through structured, machine-readable artifacts. By encoding systems security engineering techniques—such as Analytic Monitoring, Substantiated Integrity, Coordinated Defense, and Adaptive Response—into domain-level agent responsibilities, and by mapping these to AI RMF functions (Govern, Map, Measure, Manage), PBSAI offers a concrete bridge between high-level governance principles and the operational realities of AI-enabled estates.

The blueprint is intentionally tool-agnostic and incrementally adoptable. It can function as a thin overlay on an enterprise Security Operations Center (SOC) or scale to hyperscale and HPC-backed environments without altering its core structure. HPC and accelerator infrastructure expand analytic and validation capacity, but do not change the fundamental invariants of traceability, human oversight, provenance, and policy binding that define the ecosystem. Emerging quantum-assisted optimization techniques may further extend feasible evaluation and dependency analysis in high-dimensional settings, but must be integrated in ways that preserve these governance invariants.

Future work should focus on translating this architecture into reference implementations, shared schema registries, and empirical evaluation programs. Sector-specific overlays, benchmarked multi-agent testbeds, and comparative studies of governance and operational outcomes will be necessary to validate and refine the blueprint. As organizations move from isolated AI tools toward estate-level architectures, PBSAI is offered as a structured starting point for designing AI-enabled defenses that are not only powerful and scalable, but also transparent, auditable, and aligned with modern systems security engineering and AI governance expectations.

**Copyright and License**

copy, redistribute, and adapt the material in any medium or format for non-commercial purposes, provided that:

1. Attribution – Appropriate credit is given to the author, a link to the license is provided, and any changes are clearly indicated.

2. NonCommercial – The material is not used for commercial purposes.

3. ShareAlike – If you remix, transform, or build upon this work, you must distribute your contributions under the same license (CC BY-NC-SA 4.0).

No additional restrictions (such as technological protection measures or proprietary licensing) may be applied to derivatives that would prevent others from exercising the freedoms granted by this license.

For the full legal text of the license, see:
https://creativecommons.org/licenses/by-nc-sa/4.0/

Commercial use of this publication, or use under terms other than CC BY-NC-SA 4.0, requires prior written permission from the author. Nothing in this license grants rights under any patent application or patent that may cover architectural mechanisms described in this work. Commercial implementation of such mechanisms may require separate authorization.

**Funding and Acknowledgments**

This work was conducted as part of the author's role at Quantum Powered Security Inc. No external grant funding was received. The author thanks Quantum Powered Security Inc for supporting the development and publication of this work. The views expressed in this paper are those of the author and do not necessarily reflect the official positions of Quantum Powered Security Inc or its clients.

**Relationship to Prior Disclosures**

Certain architectural mechanisms referenced at a high level in this paper are the subject of U.S. provisional patent applications filed by the author. Those filings address policy-bounded AI-enabled decision mediation and multi-domain governance control for AI-enabled enterprise estates.

This paper focuses on architectural framing, threat modeling, and system-level design considerations rather than implementation details or enforcement mechanisms described in those filings. No implementation-level algorithms, synchronization protocols, or execution control logic are disclosed herein.

**Use of Artificial Intelligence Tools**

A large language model (ChatGPT, OpenAI) was used to assist with drafting, editing, and restructuring the manuscript text. The author independently determined the conceptual framing, selected and verified references, reviewed and corrected the generated text, and assumes full responsibility for the content and its accuracy.

**References**

Batool, A., Zowghi, D. and Bano, M. (2025) 'AI governance: a systematic literature review', *AI and Ethics*, advance online publication. doi:10.1007/s43681-024-00653-w. Available at: https://link.springer.com/article/10.1007/s43681-024-00653-w (Accessed: 31 January 2026).

Brundage, M. et al. (2020) 'Toward trustworthy AI development: mechanisms for supporting verifiable claims', *arXiv preprint* arXiv:2004.07213. doi:10.48550/arXiv.2004.07213. Available at: https://arxiv.org/abs/2004.07213 (Accessed: 31 January 2026).

CISA (Cybersecurity and Infrastructure Security Agency) (2020) *High Value Asset Program: Federal Civilian Executive Branch High Value Asset (HVA) Control Overlay 2.0*. Washington, DC: CISA. Available at: https://www.cisa.gov/resources-tools/resources/high-value-asset-hva-control-overlay (Accessed: 31 January 2026).

European Union (2024) *Regulation (EU) 2024/1689 of the European Parliament and of the Council of 13 June 2024 laying down harmonised rules on artificial intelligence and amending certain Union legislative acts (Artificial Intelligence Act)*. Official Journal of the European Union, L 206, pp. 1–90. Available at: https://eur-lex.europa.eu/eli/reg/2024/1689/oj/eng (Accessed: 31 January 2026).

Floridi, L. and Cowls, J. (2019) 'A unified framework of five principles for AI in society', *Harvard Data Science Review*, 1(1). doi:10.1162/99608f92.8cd550d1. Available at: https://hdsr.mitpress.mit.edu/pub/l0jsh9d1 (Accessed: 31 January 2026).

ISO/IEC (2022) *ISO/IEC 27001:2022 Information security, cybersecurity and privacy protection – Information security management systems – Requirements*. Geneva: International Organization for Standardization / International Electrotechnical Commission. Available at: https://www.iso.org/standard/82875.html (Accessed: 31 January 2026).

ISO/IEC (2023) *ISO/IEC 42001:2023 Information technology – Artificial intelligence – Management system*. Geneva: International Organization for Standardization / International Electrotechnical Commission. Available at: https://www.iso.org/standard/81230.html (Accessed: 31 January 2026).

Jobin, A., Ienca, M. and Vayena, E. (2019) 'The global landscape of AI ethics guidelines', *Nature Machine Intelligence*, 1(9), pp. 389–399. doi:10.1038/s42256-019-0088-2. Available at: https://www.nature.com/articles/s42256-019-0088-2 (Accessed: 31 January 2026).

NIST (National Institute of Standards and Technology) (2020) *Security and privacy controls for information systems and organizations* (NIST Special Publication 800-53,

Revision 5). Gaithersburg, MD: U.S. Department of Commerce. Available at: https://csrc.nist.gov/publications/detail/sp/800-53/rev-5/final (Accessed: 31 January 2026).

NIST (National Institute of Standards and Technology) (2021) *Developing cyber-resilient systems: a systems security engineering approach* (NIST Special Publication 800-160, Volume 2, Revision 1). Gaithersburg, MD: U.S. Department of Commerce. Available at: https://csrc.nist.gov/publications/detail/sp/800-160/vol-2-rev-1/final (Accessed: 31 January 2026).

NIST (National Institute of Standards and Technology) (2023) *Artificial Intelligence Risk Management Framework (AI RMF 1.0)*. Gaithersburg, MD: U.S. Department of Commerce. Available at: https://nvlpubs.nist.gov/nistpubs/ai/NIST.AI.100-1.pdf (Accessed: 31 January 2026).

OECD (Organisation for Economic Co-operation and Development) (2019) *Recommendation of the Council on Artificial Intelligence* (OECD/LEGAL/0449). Paris: OECD. Available at: https://legalinstruments.oecd.org/en/instruments/OECD-LEGAL-0449 (Accessed: 31 January 2026).

OMB (Office of Management and Budget) (2018) *Memorandum M-19-03: Strengthening the Cybersecurity of Federal Agencies by Enhancing the High Value Asset Program*. Washington, DC: Executive Office of the President. Available at: https://www.whitehouse.gov/wp-content/uploads/2018/12/M-19-03.pdf (Accessed: 31 January 2026).

Papagiannidis, E., Mikalef, P. and Conboy, K. (2025) 'Responsible artificial intelligence governance: a review and research framework', *The Journal of Strategic Information Systems*, 34(2), p. 101885. doi:10.1016/j.jsis.2024.101885. Available at: https://www.sciencedirect.com/science/article/pii/S0963868724000672 (Accessed: 31 January 2026).

Sommer, R. and Paxson, V. (2010) 'Outside the closed world: on using machine learning for network intrusion detection', in *2010 IEEE Symposium on Security and Privacy*. Oakland, CA: IEEE, pp. 305–316. doi:10.1109/SP.2010.25. Available at: https://doi.org/10.1109/SP.2010.25 (Accessed: 31 January 2026).

Tounsi, W. and Rais, H. (2018) 'A survey on technical threat intelligence in the age of sophisticated cyber attacks', *Computers & Security*, 72, pp. 212–233. doi:10.1016/j.cose.2017.09.001. Available at: https://www.sciencedirect.com/science/article/pii/S0167404817301839 (Accessed: 31 January 2026).

Weidinger, L. et al. (2022) 'Taxonomy of risks posed by language models', in *Proceedings of the 2022 ACM Conference on Fairness, Accountability, and Transparency (FAccT '22)*. Seoul, Republic of Korea: ACM, pp. 214–229.

doi:10.1145/3531146.3533088. Available at: https://dl.acm.org/doi/10.1145/3531146.3533088 (Accessed: 31 January 2026).

**Appendix A - Master Agent Index (Domains A–L)**

| Domain | Agent ID | Agent Name |
|---|---|---|
| **A – GRC & Oversight** | A1 | Governance Policy Agent |
| | A2 | Risk Analysis Agent |
| | A3 | Compliance Auditor Agent |
| | A4 | Policy Authoring Agent |
| | A5 | Evidence Collector Agent |
| | A6 | Control Monitor Agent |
| | A7 | Metrics Aggregator Agent |
| | A8 | Supplier Risk Agent |
| **B – Asset, Configuration & Change Management** | B1 | Asset Discovery Agent |
| | B2 | Baseline Compliance Agent |
| | B3 | Config Monitor Agent |
| | B4 | Change Governance Agent |
| | B5 | Patch Orchestration Agent |
| | B6 | Vulnerability Scanning Agent |
| | B7 | Risk Prioritization Agent |
| **C – Identity, Credential & Access Management (ICAM)** | C1 | Identity Lifecycle Agent |
| | C2 | Credential Issuance Agent |
| | C3 | Access Review Agent |
| | C4 | Privileged Access Agent |
| | C5 | Federation Trust Agent |

| Domain | Agent ID | Agent Name |
|---|---|---|
| | C6 | Policy Enforcement Agent |
| | C7 | Session Monitoring Agent |
| **D – Threat Intelligence & Monitoring** | D1 | Threat Intel Aggregator Agent |
| | D2 | SIEM Analyst Agent |
| | D3 | UEBA Analysis Agent |
| | D4 | Autonomous Hunter Agent |
| | D5 | Exposure Mapping Agent (ASM) |
| | D6 | Honeypot Controller Agent |
| **E – Protect & Harden** | E1 | Endpoint Protection Agent |
| | E2 | Network Defense Agent |
| | E3 | Data Protection Agent |
| | E4 | Application Security Agent |
| | E5 | Encryption Management Agent |
| | E6 | DLP Enforcement Agent |
| | E7 | Secure Configuration Agent |
| **F – Data Security & Privacy** | F1 | Data Classification Agent |
| | F2 | Data Access Governance Agent |
| | F3 | Privacy Compliance Agent |
| | F4 | Data Retention Agent |
| | F5 | Data Masking Agent |
| | F6 | Cross-Border Transfer Agent |

| Domain | Agent ID | Agent Name |
|---|---|---|
| | F7 | Consent Management Agent |
| **G – Incident Response & Digital Forensics (IR/DFIR)** | G1 | Incident Triage Agent |
| | G2 | Automated Containment Agent |
| | G3 | Recovery Orchestration Agent |
| | G4 | Forensic Analysis Agent |
| | G5 | RCA Analysis Agent |
| | G6 | Communications Agent |
| **H – Resilience & Business Continuity** | H1 | Business Continuity Planner Agent |
| | H2 | Disaster Recovery Orchestrator Agent |
| | H3 | Backup Validation Agent |
| | H4 | Crisis Simulation Agent |
| | H5 | Dependency Mapping Agent |
| | H6 | Resilience Metrics Agent |
| **I – Security Architecture & Systems Engineering** | I1 | Architecture Advisor Agent |
| | I2 | SSE Process Agent |
| | I3 | Crypto Architecture Agent |
| | I4 | Protocol Compliance Agent |
| | I5 | Pattern Management Agent |
| | I6 | DevSecOps Pipeline Agent |
| **J – Physical & Environmental Security** | J1 | Facility Access Agent |

| Domain | Agent ID | Agent Name |
|---|---|---|
| | J2 | Environmental Monitoring Agent |
| | J3 | Asset Protection Agent |
| | J4 | Physical Incident Agent |
| | J5 | Visitor Management Agent |
| | J6 | Safety Compliance Agent |
| K – Supply Chain & Lifecycle Security | K1 | Procurement Policy Agent |
| | K2 | Authenticity Verification Agent |
| | K3 | SBOM Analysis Agent |
| | K4 | Secure Disposal Agent |
| L – Security Program Enablement | L1 | Security Knowledge Agent |
| | L2 | Security Analytics Agent |
| | L3 | SOAR Orchestration Agent |
| | L4 | Data Curation Agent |
| | L5 | AI Validation Agent |

## Appendix B – Example Agent Specification: C1 Identity Provisioning (Domain C – ICAM)

This appendix provides a condensed, worked example of a PBSAI agent specification. The C1 Identity Provisioning Agent lives in Domain C – Identity, Credential, and Access Management (ICAM) and automates joiner–mover–leaver (JML) provisioning and deprovisioning across directories and applications. It illustrates how PBSAI agents use MCP-style context envelopes, Output Contracts, state machines, and SLOs in practice. A complete, extended version of this specification (including additional contracts, feedback loops, and protocol details) is provided in the companion materials.

### B.1 Purpose and scope

**Purpose.**
C1 automates identity provisioning for employees and contractors, with a focus on:

- Joiners: creating accounts and applying baseline roles/entitlements.
- Movers: adjusting entitlements when roles, departments, or locations change.
- Leavers: disabling and deprovisioning accounts and credentials promptly.

**Scope.**

- Primary targets: enterprise IdP(s) (for example, Azure AD, Okta), core SaaS applications, and key internal platforms.
- Policy goals: least privilege, separation of duties (SoD), just-in-time / just-enough access, and timely deprovisioning.

C1 is event-driven by HRIS and identity events and operates under MCP-style envelopes and governance constraints defined in Domain A.

### B.2 Inputs and outputs (high-level blueprint)

**Authoritative inputs**

- **HRIS events** (webhook/REST)
    - Event types: hire, transfer, terminate, extended_leave, return_from_leave.
    - Fields: employee_id, name, department, job_title, location, manager_id, employment_type, start_date, end_date.
- **Role Model Catalog** (Git/REST)
    - Mapping from role_id → entitlements (groups, app roles, permissions).
    - SoD conflict rules and approver chains.

- **Identity Graph** (Domain C store / graph)
    - Current accounts, groups, roles, and access history per identity.
- **Policies and constraints** from Domain A
    - Account management policy, SoD policy, emergency access policy, JML SLAs.

**Outputs (summary)**

C1 primarily emits:

- **SCIM mutations** to identity providers and directories (create/update/deprovision).
- **Access change notifications** for downstream systems (PAM, data platforms, audit lake).
- **SoD violation / exception records** when conflicts cannot be auto-resolved.
- **Feedback events** (success/failure, contestation) for learning and validation.

These outputs are expressed as Output Contracts (OCs) and wrapped in MCP-style envelopes as described in Section 4 of the main text.

**B.3 Example MCP-style input envelope**

When C1 is triggered by a new hire event, an orchestrator in Domain L constructs an MCP-style envelope and passes it to the agent with the HRIS payload:

```
{
  "mission_id": "mission-joiners-q1",
  "thread_id": "thread-emp-78421",
  "task_id": "task-c1-provision-78421",
  "role": "IdentityProvisioningAgent",
  "intent": "provision_joiner",
  "policy_refs": ["AC-2", "IA-2", "SoD-Policy-v3"],
  "constraints": [
   "no_emergency_admin",
   "HITL_on_SoD_conflict",
   "timebox_15m"
  ],
```

```
  "decision_basis": {
    "evidence_refs": [
      "uri://hris/events/hris-evt-78421",
      "uri://role-catalog/snapshot/2025-11-01"
    ],
    "confidence": 0.0
  },
  "provenance": {
    "producer_spiffe": "spiffe://enterprise/orch/icam",
    "signing_kid": "kid-orch-icam-01",
    "attestation_ref": "uri://slsa/attestations/orch-icam@sha256:abcd..."
  },
  "classification": "internal-plus-sensitive",
  "legal_hold": false
}
```

C1 consumes this envelope plus the HRIS event, applies deterministic and model-assisted logic, then emits one or more Output Contracts with updated envelope fields (for example, confidence, evidence_refs).

**B.4 Example Output Contract: OC-1 SCIMMutation**

This example shows how a single Output Contract is specified. In the full specification there are additional contracts (for access-change notifications, SoD violations, feedback, and so on); here we focus on the primary provisioning action.

**OC-1: SCIMMutation**

- **Producer(s):** C1 Identity Provisioning Agent
- **Purpose:** Express create, update, and deprovision operations for user or group identities in a SCIM-compatible directory.
- **Transport / Topic:** HTTPS SCIM endpoint per target (for example, /scim/v2/Users, /scim/v2/Groups).

**Schema (simplified JSON shape).**

```
{
```

```json
"oc_type": "SCIMMutation",
"mutation_id": "uuid-1234-5678",
"target_system": "azure_ad",
"operation_type": "create | update | deprovision",
"target_resource": "user | group",
"target_id": "user-or-group-id-or-null-for-create",
"employee_id": "78421",
"scim_payload": { "...": "..." },
"reason": "joiner | mover | leaver | remediation",
"requested_at": "2025-11-01T12:00:00Z",
"effective_ts": "2025-11-01T12:05:00Z",
"idempotency_key": "sha256(target_system, operation_type, employee_id, floor_60s)",
"trace_id": "trace-xyz",
"mission_id": "mission-joiners-q1",
"thread_id": "thread-emp-78421"
}
```

**Key semantics.**

- **Delivery:** exactly-once per combination of (target_system, operation_type, employee_id, version_floor_60s) (consumer enforces idempotency).

- **Ordering:** per employee_id.

- **Ack/Nack:** HTTP 2xx → ACK; 4xx/5xx → NACK → retry with exponential backoff up to a bounded count, then open an incident/ticket.

- **Side effects:** update identity graph; emit a separate AccessChangeEvent (OC-2) for downstream consumers.

- **HITL:** when SoD conflicts remain unresolved or privileged roles are involved, mutations are emitted in "pending" state and require explicit human co-sign, referenced via MCP policy_cosign constraints.

- **Retention:** sufficient journal of request/response hashes for audit (for example, 7 years), but with PII minimization where possible.

The combination of a strongly typed OC and a signed MCP envelope gives downstream systems both machine-checkable structure and governance context for each provisioning action.

### B.5 State machine snapshot

C1 follows a simple state machine for each JML task. Below is a concise view of the main states and transitions:

**States.**

- idle
- awaiting_hris_event
- role_resolution
- sod_check
- awaiting_approval
- provisioning
- verifying
- sla_breach
- contest_workflow
- rolled_back
- closed

**Key transitions.**

- idle → awaiting_hris_event when a new HRIS JML event arrives.
- awaiting_hris_event → role_resolution when the event is valid and effective_date ≤ now.
- role_resolution → sod_check when baseline roles have been derived from HRIS attributes.
- sod_check → awaiting_approval when a separation-of-duties conflict is detected and cannot be auto-resolved.
- sod_check → provisioning when no SoD conflicts are found.
- awaiting_approval → provisioning when required approvers have co-signed.
- provisioning → verifying when SCIMMutation calls are ACKed by target systems.
- verifying → closed when observed entitlements match the desired state.

- provisioning → rolled_back on partial failure where automated rollback is possible.

- * → sla_breach when elapsed time exceeds the defined JML SLA for the action (for example, joiner or leaver beyond the allowed window).

- * → contest_workflow when an AccessDispute or SoDViolation is raised (for example, a user contests an access change).

Time-based and retry behaviors (for example, retry limits, backoff, circuit breakers) are defined per OC and interpreted under this state machine. All transitions and decisions inherit the MCP envelope, which captures mission, policy, constraints, and provenance.

### B.6 SLOs and KPIs

C1 is governed by a small set of **service level objectives (SLOs)** and **key performance indicators (KPIs)** that tie technical behavior to governance expectations:

- **JoinerProvisioning_p95 ≤ 15 minutes**
    - 95% of joiner events (hire or return-from-leave) result in required baseline access being in place within 15 minutes of the effective time.

- **LeaverDisable_p95 ≤ 5 minutes**
    - 95% of leaver events (termination or leave) result in account disablement within 5 minutes of the effective time.

- **MoverDelta_p95 ≤ 10 minutes**
    - 95% of role/location/department changes result in entitlements being updated within 10 minutes.

- **SoDFalseBlockRate_30d ≤ 2%**
    - Over a 30-day window, fewer than 2% of SoD-based blocks are later judged to have been unjustified.

- **RollbackSuccessRate_7d ≥ 99%**
    - Over a 7-day window, at least 99% of triggered rollbacks complete successfully and restore prior access state.

These SLOs are tracked via Domain L analytics agents using the same Output Contracts (for example, SCIMMutation, AccessChangeEvent, IncidentTimeline). Results feed back to Domain A for risk reporting and policy refinement.

**B.7 Relationship to the broader ecosystem**

This example shows how a single PBSAI agent:

- Implements the agent pattern from Section 4 (deterministic-first, bounded LLM use, MCP envelope, Output Contracts).

- Anchors its behavior in identity governance and SoD policies from Domain A.

- Interacts with other domains (B for assets, D for monitoring, G for incidents, L for orchestration and analytics).

In practice, C1 is one of several Domain C agents (for example, identity proofing, credential lifecycle, privileged access brokers) that, together with agents in other domains, form the multi-agent governance ecosystem described in the main text.